\documentclass[journal]{IEEEtran}

\usepackage{amsmath,amssymb,amsfonts}
\usepackage{cite,cleveref,enumitem}
\usepackage{bbm,amsthm}
\usepackage{algorithm,algorithmic}
\usepackage{textcomp}
\usepackage{graphicx,xcolor}
\usepackage{multirow}
\usepackage[english]{babel}
\usepackage{caption}
\usepackage{subcaption}	
\graphicspath{{Figures/}}
\usepackage{ifpdf}
\ifpdf
  \usepackage{epstopdf}
\fi

\crefname{figure}{fig.}{figs.}
\crefname{section}{sec.}{secs.}

\newcommand{\upi}{\Upsilon_{\mathrm{upper}}}

\crefname{app}{Appendix}{Appendices}
\crefname{cor}{Corollary}{Corollary}
\crefname{prop}{Proposition}{Proposition}
\crefname{lemma}{Lemma}{Lemma}
\crefname{defn}{Definition}{Definition}
\crefname{conj}{Conjecture}{Conjecture}
\crefname{exam}{Example}{Example}
\crefname{supp}{Supplemental Section}{Supplemental Section}

\newcommand{\bs}{\boldsymbol}
\newcommand{\bb}{\mathbb}
\newcommand{\mcal}{\mathcal}

\newcommand{\one}{\bs{1}}

\newcommand{\lb}{\left(}
\newcommand{\rb}{\right)}
\newcommand{\ls}{\left[}
\newcommand{\rs}{\right]}
\newcommand{\lc}{\left\{}
\newcommand{\rc}{\right\}}

\newcommand{\lv}{\left\vert}
\newcommand{\rv}{\right\vert}

\newcommand{\LRV}[1]{{\left\vert\kern-0.25ex\left\vert\kern-0.25ex\left\vert #1 \right\vert\kern-0.25ex\right\vert\kern-0.25ex\right\vert}}

\newcommand{\expect}[2]{\bb{E}_{#1}\lc#2\rc}

\newcommand{\nth}{^\mathsf{th}}

\newcommand{\bbP}{\bb{P}}

\newcommand{\calA}{\mcal{A}}

\newcommand{\calN}{\mcal{N}}
\newcommand{\calO}{\mcal{O}}

\newcommand{\vecs}{\bs{s}}

\newcommand{\vecy}{\bs{y}}

\newcommand{\vecmu}{\bs{\mu}}
\newcommand{\vecnu}{\bs{\nu}}

\newcommand{\vecsigma}{\bs{\sigma}}

\begin{document}

\title{Scalable and Decentralized Algorithms for Anomaly Detection via Learning-Based Controlled Sensing
}

\author{  Geethu~Joseph, Chen~Zhong,
       M.~Cenk~Gursoy, Senem~Velipasalar, and Pramod~K.~Varshney,~\IEEEmembership{Life~Fellow,~IEEE}
\thanks{
This work was presented in part at the IEEE International Conference on Communications, June 2021, Montreal, Canada.

G. Joseph is with the Department of Microelectronics, Delft University of Technology, Delft 2628, Netherlands.
Email: g.joseph@tudelft.nl

C. Zhong, M. C. Gursoy, S. Velipasalar, and P. K. Varshney are with the Department
of Electrical and Computer Engineering, Syracuse University, New York, 13244, USA. Emails:\{czhong03,mcgursoy,svelipas,varshney\}@syr.edu

}
}


\maketitle

\begin{abstract}
We address the problem of sequentially selecting and observing processes from a given set to find the anomalies among them. The decision-maker observes a subset of the processes at any given time instant and obtains a noisy binary indicator of whether or not the corresponding process is anomalous. In this setting, we develop an anomaly detection algorithm that chooses the processes to be observed at a given time instant, decides when to stop taking observations, and declares the decision on anomalous processes. The objective of the detection algorithm is to identify the anomalies with an accuracy exceeding the desired value while minimizing the delay in decision making. We devise a centralized algorithm where the processes are jointly selected by a common agent as well as  a decentralized algorithm where the decision of whether to select a process is made independently for each process. Our algorithms rely on a Markov decision process defined using the marginal probability of each process being normal or anomalous, conditioned on the observations. We implement the detection algorithms using the deep actor-critic reinforcement learning framework.  Unlike prior work on this topic that has exponential complexity in the number of processes, our algorithms have computational and memory requirements that are both polynomial in the number of processes. We demonstrate the efficacy of these algorithms using numerical experiments by comparing them with state-of-the-art methods.
\end{abstract}


\IEEEpeerreviewmaketitle

\section{Introduction}
We consider the problem of observing a given set of processes to detect the anomalies among them via controlled sensing. Here, the decision-maker does not observe all the processes at each time instant, but sequentially selects and observes a small subset of processes at a time. The sequential control of the observation process is referred to as controlled sensing. The challenge here is to devise a selection policy to sequentially choose the processes to be observed so that the decision is accurate and fast. This problem arises, for instance, in sensor networks used for remote health monitoring, structural health monitoring, etc~\cite{chung2006remote,bujnowski2013enhanced}. Such systems are equipped with different types of sensors to monitor different functionalities (or processes) of the system.  The sensors send their observations to a common decision-maker that identifies any potential system malfunction. These sensor observations can be noisy due to faulty hardware or unreliable communication links. Therefore, to ensure the accuracy of the decision, we employ a sequential process selection strategy that observes a subset of processes over multiple time instants before the final decision is made. Further, the different processes can be statistically dependent on each other, and as a result, observing one process also gives information about the other dependent processes. Our goal is to derive a selection policy that accurately identifies the anomalous processes with minimum delay by exploiting the underlying statistical dependence among the processes.

\subsection{Related Literature}
Anomaly detection problem is a well-studied research topic, and there are several active sensing schemes for anomaly detection designed to monitor the environments~\cite{6815791,zhu2021cost,lai2011quickest,hemo2020searching}. A popular approach for solving the anomaly detection problem is to use the active hypothesis testing framework~\cite{zhong2019deep,joseph2020anomaly}. Here,  the decision-maker defines a hypothesis corresponding to each of the possible states of the processes and computes the posterior probabilities over the hypothesis set using the observations. The decision-maker continues to collect observations until the probability corresponding to one of the hypotheses exceeds the desired confidence level. This framework of active hypothesis testing was introduced by Chernoff in \cite{chernoff1959sequential} which dates back to 1959. This seminal work also established the asymptotical optimality of the test for binary hypotheses.  Later, the asymptotic optimality of the test was extended to the multiple hypothesis testing problem~\cite{nitinawarat2013controlled}. Following the Chernoff test, several other model-based tests were also studied in the literature~\cite{bessler1960theory,naghshvar2013active,franceschetti2016chernoff,huang2018active}. While the detection accuracy of these algorithms are of significant research interest, several studies focus on the sensing costs and/or switching costs during the detection process. For instance, in~\cite{huang2018active,gurevich2019sequential,chen2019active,qin2021low,qin2021active}, the authors seek to minimize either one type of cost or both costs jointly.

Most of the above works focus on centralized algorithms where the processes are selected jointly by a common agent. A few other works discuss the model-based non-adaptive detection algorithms in a decentralized setting~\cite{blum1997distributed,viswanathan1997distributed,mei2008asymptotic,wang2011asymptotic,
chair1988distributed,pados1995distributed}.  Recently, a model-based active hypothesis testing algorithm was also studied in the literature~\cite{rangi2018decentralized}. However, this work considered the case wherein each sensor in the decentralized network estimated the true hypothesis independently, and the network arrived at a consensus based on the sensor decisions and associated confidence levels. This framework does not address the case wherein each sensor in the network can only observe the corresponding process, and they cooperate to estimate the states of all the processes.

Recently, the active hypothesis testing framework has been combined with deep learning algorithms to design data-driven anomaly detection algorithms~\cite{kartik2018policy,zhong2019deep,joseph2020anomaly,joseph2020anomaly2}. These algorithms learn from a training dataset and provide the added advantage of adaptability to the underlying statistical dependence among the processes. The state-of-the-art algorithms in this direction employ the reinforcement learning (RL) algorithms such as Q-learning~\cite{kartik2018policy} and actor-critic~\cite{zhong2019deep,joseph2020anomaly}; and the active inference framework~\cite{joseph2020anomaly2}. However, the major drawback of this solution strategy is the heavy computational burden that arises due to a large number of hypotheses. Since each process can either be normal or anomalous (two states per process), the number of hypotheses increases exponentially with the number of processes. Hence, our paper focuses on the design of a scalable anomaly detection algorithm for anomaly detection via learning-based controlled sensing, whose complexity is polynomial in the number of processes. 

Moreover, the above existing literature on deep active hypothesis testing thus far focuses on centralized algorithms where the processes are selected jointly by a common agent. As we mentioned above, centralized algorithms do not scale with the number of processes. Also, the central decision-making agent introduces a single point of failure, thus rendering this architecture is not suitable for monitoring critical applications. So to handle this problem, we also explore a decentralized algorithm for anomaly detection. In the decentralized version, there is no common agent that collects the observations and makes the process selection decisions. Instead, each sensor independently decides whether or not to observe their corresponding processes. In short, in this paper, we attempt to devise a learning-based controlled sensing framework for anomaly detection with polynomial complexity in the number of processes for two settings: the centralized and decentralized cases.

\subsection{Our Contributions}
The specific contributions of the paper are as follows:
\begin{itemize}
\item \emph{Estimation algorithm:}  In \Cref{sec:estimation}, we first reformulate the problem of anomaly detection in terms of the marginal (not joint) probability of each process being normal or anomalous, conditioned on the observations. Consequently, the number of posterior probabilities computed by the algorithm at every time instant is linear in the number of processes. Based on these marginal posterior probabilities, we define the notion of a confidence level that is proportional to the decision accuracy. We then derive an algorithm in \Cref{sec:estimation} for estimating the true states of the processes when the confidence level exceeds a predefined threshold.
\item \emph{Centralized algorithm:} In \Cref{sec:central}, we present a novel centralized algorithm for anomaly detection where we restrict the number of processes to be chosen at any time to be one. To obtain the algorithm, we define two reward functions that monotonically increase with the decision accuracy and decrease with the duration of the observation acquisition phase. These definitions allow us to reformulate the anomaly detection problem as a long-term average reward maximization problem in a Markov decision process (MDP) where the MDP state is the marginal probability vector. This problem is solved using a policy gradient RL algorithm called the actor-critic method, and the algorithm is implemented using deep neural networks.
\item \emph{Decentralized algorithm:} In \Cref{sec:decentral}, we propose a decentralized version of our centralized algorithm. Here, at each time instant, the selection decision is independently made for each process, and as a consequence, we allow the algorithm to choose more than one process at a time. The number of observations is reduced by modifying the notion of reward to accommodate the sensing cost. Using the modified notion of MDP, we derive a novel actor-critic algorithm for decentralized anomaly detection.
\item \emph{Empirical results:} Using numerical results, we compare our algorithms to the state-of-the-art algorithms in \Cref{sec:simulations}. We show that the centralized algorithm can learn and adapt to the statistical dependence among the processes. The decentralized algorithm also offers a good accuracy level in detecting anomalies.
\end{itemize}

Overall, this paper presents centralized and decentralized algorithms for anomaly detection.  The polynomial complexity of the algorithms makes them scalable, and thus, practically more useful.

Furthermore, this journal paper makes several new contributions compared to the conference version~\cite{joseph2020scalable}. In addition to the entropy-based reward function in the conference version, we also present a log-likelihood ratio (LLR)-based reward function for the scalable algorithms (see \Cref{sec:central}). This scheme is empirically shown to slightly outperform the entropy-based scheme. We further introduce the concept of decentralized anomaly detection and present a new actor-critic algorithm based on the concept of \emph{centralized training and decentralized execution} (see \Cref{sec:decentral}). Also, we provide a detailed derivation of recursive updates for the marginal probabilities (see \Cref{sec:estimation}) and the pairwise probabilities (see \Cref{sec:simulations}).

\vspace{0.2cm}
\noindent\emph{Organization:} The remainder of the paper is organized as follows. We present the system model for the anomaly detection problem in \Cref{sec:system}. \Cref{sec:estimation} describes the estimation algorithm that is common to both centralized and decentralized algorithms.  In \Cref{sec:central,sec:decentral}, we present our centralized and decentralized selection policies, respectively. We provide the simulation results in \Cref{sec:simulations} and offer our concluding
remarks in \Cref{sec:conclusion}.

\section{Anomaly Detection Problem}\label{sec:system}
We consider a set of $N$ processes where the state of each process is a binary random variable. The process state vector is denoted by $\vecs\in\{0,1\}^N$ whose $i\nth$ entry $\vecs_i$ being $0$ and $1$ indicates that the $i\nth$ process is in the normal state and the anomalous state, respectively. We aim to detect the anomalous processes, which is equivalent to estimating the random binary vector $\vecs$.

Each process is monitored by a sensor. We estimate the process state vector $\vecs$ by selecting and observing a subset of the processes at every time instant. Let $\calA(k)\subseteq\{1,2,\ldots,N\}$ be the set of processes observed by the algorithm at time instant $k$, and the corresponding observation be $\vecy_{\calA(k)}(k)\in\{0,1\}^{\lv\calA(k)\rv}$. We assume that an observation corresponding to a process is noisy and has a fixed probability of being erroneous. Specifically, the uncertainty in the observation is modeled using the following probabilistic model for any $i\in\{1,2,\ldots,N\}$,
\begin{equation}\label{eq:mesurement}
\vecy_{i}(k) = \begin{cases}
 \vecs_{i} & \text{ with probability } 1-p,\\
1-\vecs_{i}& \text{ with probability } p,
\end{cases}
\end{equation}
where $p\in[0,1]$ is called the flipping probability. Further, we assume that conditioned on the value of $\vecs$, the observations obtained across different time
instants are jointly (conditionally) independent, i.e., for any $k$,
\begin{equation}
\bbP\ls \lc \vecy_{i}(l), i=1,2,\ldots,N\rc_{l=1}^k\middle|\vecs\rs = \prod_{i=1}^N\prod_{l=1}^k\bbP\ls  \vecy_{i}(l)\middle|\vecs_i\rs.\label{eq:indep}
\end{equation}
Therefore, the observations corresponding to the $i\nth$ process $\lc\vecy_i(k)\in\{0,1\}\rc_{k=1}^{\infty}$ is a sequence of independent and identically distributed binary random variables parameterized by the process state $\vecs_i\in\{0,1\}$.

At each time instant, the decision-maker computes an estimate of $\vecs$ along with the confidence in the estimate using $\vecy_{\calA(k)}(k)$. The decision-maker observes the processes until the confidence exceeds the desired level denoted by $\upi\in(0,1)$. In short, we have two interrelated tasks: one, to develop an algorithm to estimate the process state vector and the associated confidence in the estimate; and two, to derive a policy that decides the processes to be observed at each time instant and the criterion to stop collecting observations.  We seek the estimation algorithm and the policy that jointly minimize the stopping time $K$ while ensuring a desired accuracy level. Here, the stopping time refers to the time instant at which the observation acquisition phase ends.

Our goal is to develop scalable algorithms that have polynomial time complexity in $N$, and we consider two settings: centralized and decentralized algorithms. In the centralized setting, the algorithm selects the processes to be probed using a centralized agent, whereas, in the decentralized setting, the selection decision is made independently across the processes. 

We next present our estimation algorithm that is common to both centralized and decentralized algorithms.

\section{Estimation Algorithm}\label{sec:estimation}
In this section, we derive an algorithm to estimate the process state vector from the observations. We note that the observations depend on the selection policy, and the policy design, in turn, depends on the estimation algorithm. We first present the estimation algorithm and then derive selection policies based on the estimation objectives in the next sections.

To estimate the process state vector, we first compute the belief vector $\vecsigma(k)\in[0,1]^N$ at time $k$ whose $i\nth$ entry $\vecsigma_i(k)$ is the posterior probability that the $i\nth$ process is normal ($\vecs_i=0$). So, the probability that the $i\nth$ process is anomalous ($\vecs_i=1$) is $1-\vecsigma_i(k)$. As each observation arrives, we update the  belief vector as follows:
\begin{align}
\vecsigma_i(k) &= \bbP\ls \vecs_i= 0 \middle| \lc \vecy_{\calA(l)}(l)\rc_{l=1}^{k} \rs\label{eq:posterior_01}\\
&=\frac{\bbP\ls \lc \vecy_{\calA(l)}(l)\rc_{l=1}^{k}\middle| \vecs_i= 0 \rs\bbP\ls \vecs_i=0\rs}{\displaystyle\sum_{s=0,1}\bbP\ls \lc \vecy_{\calA(l)}(l)\rc_{l=1}^{k}\middle| \vecs_i= s \rs\bbP\ls \vecs_i=s\rs}.\label{eq:posterior_1}
\end{align}
Here, we approximate the joint probability distribution by assuming that the observations at time $k$ are independent of each other and the past observations, conditioned on the process state $\vecs_i$:
\begin{multline}
\bbP\ls \lc \vecy_{\calA(l)}(l)\rc_{l=1}^{k}\middle| \vecs_i= s \rs
 \approx \displaystyle \bbP\ls \lc \vecy_{\calA(l)}(l)\rc_{l=1}^{k-1} \middle| \vecs_i= s \rs\\\times\prod_{a\in\calA(k)}\bbP \ls \vecy_{a}(k) \middle| \vecs_i= s \rs. \label{eq:approx}
\end{multline}
From \eqref{eq:indep}, the observation $\vecy_{\calA(k)}(k)$ is independent of all other observations, conditioned on the value of $\vecs_{\calA(k)}$.
As a result, the approximation is exact when $\vecs_{\calA(k)}$ is a deterministic function of $\vecs_i$. From \eqref{eq:approx}, we arrive at
\begin{multline*}
\bbP\ls \lc \vecy_{\calA(l)}(l)\rc_{l=1}^{k}\middle| \vecs_i= s \rs\bbP\ls \vecs_i=s\rs\\
\approx \bbP\ls \vecs_i=s \middle|\lc  \vecy_{\calA(l)}(l)\rc_{l=1}^{k-1} \rs\bbP \ls \lc  \vecy_{\calA(l)}(l)\rc_{l=1}^{k-1}  \rs\\
\times \prod_{a\in\calA(k)}\bbP \ls \vecy_{a}(k) \middle| \vecs_i= s \rs.
\end{multline*}
Substituting the above relation into \eqref{eq:posterior_1}, we obtain the recursive formula,
\begin{equation}
\vecsigma_i(k)\approx  \frac{\vecsigma_i(k-1)\prod_{a\in\calA(k)}\bbP \ls \vecy_{a}(k) \middle| \vecs_i= 0 \rs}{
\displaystyle\sum_{s=0,1}\lv s-\vecsigma_i(k-1)\rv\prod_{a\in\calA(k)}\bbP \ls \vecy_{a}(k) \middle| \vecs_i= s \rs},\label{eq:sigma_update1}
\end{equation}
because $\bbP\ls \vecs_i=1 \middle|\lc  \vecy_{\calA(l)}(l)\rc_{l=1}^{k-1} \rs= 1-\vecsigma_i(k-1)$ from \eqref{eq:posterior_01}.
Further, the conditional probability $\bbP \ls \vecy_{a}(k) \middle| \vecs_i= s \rs$ for $s=0,1$ is given by
\begin{align}
\bbP \ls \vecy_{a}(k) \middle| \vecs_i= s \rs\notag\\
&\hspace{-1.9cm}=\displaystyle \sum_{s'=0,1}\bbP \ls \vecy_{a}(k) \middle| \vecs_{a}= s' \rs\bbP\ls \vecs_{a} =s'\middle| \vecs_i=s\rs\\
&\hspace{-1.9cm}=\displaystyle  \sum_{s'=0,1} p^{\lv s'- \vecy_{a}(k)\rv }(1-p)^{\lv 1-s'- \vecy_{a}(k)\rv } \bbP\ls \vecs_{a} =s'\middle| \vecs_i=s\rs,\label{eq:sigma_update3}
\end{align}
which follows from \eqref{eq:mesurement}. Here, the term $\bbP\ls \vecs_{a} =s'\middle| \vecs_i=s\rs$, which depends on the statistical dependence between the processes, can be either assumed to be known or easily estimated from the training data\footnote{During the training phase, the true value of $\vecs$ is provided, but the optimal selection at each time instant is unknown.}
for every pair $(i,j)$.  Thus, \eqref{eq:sigma_update1} and \eqref{eq:sigma_update3} give the recursive update of $\vecsigma(k)$.

We note that when $\vecs_{\calA(k)}$ and $\vecs_i$ are statistically independent, for $s=0,1$ and $a\in\calA(k)$,
\begin{equation*}
\bbP \ls \vecy_{a}(k) \middle| \vecs_i= s \rs =  \bbP \ls \vecy_{a}(k) \rs.
\end{equation*}
Consequently, \eqref{eq:sigma_update1} reduces to $\vecsigma_i(k)=\vecsigma_i(k-1)$. This update is intuitive since an observation from process $\vecs_{\calA(k)}$ does not change the probabilities associated with an independent process $\vecs_i$. In other words, the recursive relation is exact when $\vecs_i$ and $\vecs_{\calA(k)}$ are either independent or $\vecs_{\calA(k)}$ is a deterministic function of  $\vecs_i$.  We discuss this point in detail in \Cref{sec:simulations}.

Once $\vecsigma(k)$ is obtained,  the $i\nth$ component of the process state vector estimate denoted by $\hat{\vecs}(k)$ can be determined in a straightforward fashion as
\begin{equation}\label{eq:estimate}
\hat{\vecs}_i = \begin{cases}
0 & \text{ if } \vecsigma_i(k)\geq 1-\vecsigma_i(k)\\
1 & \text{ if } \vecsigma_i(k)< 1-\vecsigma_i(k).
\end{cases}
\end{equation}

Hence, the derivation of the estimation algorithm is complete and we next discuss the design of the selection policy. The design of the selection policy (i.e., the policy to determine which processes to observe at a given time) is a sequential decision-making problem, and this problem can be formulated using the mathematical framework of MDPs. The MDP-based formulation allows us to obtain the selection policy via reward maximization of the MDP using RL algorithms. In the following sections, we present the MDP and RL algorithms for the centralized and decentralized settings.

\section{Scalable Centralized Decision-Making and Anomaly Detection}\label{sec:central}
In the centralized version, we restrict $\lv\calA(k)\rv=1$ for all values of $k$ for simplicity, i.e., only one process is observed per time instant.
The observation obtained by a sensor at a given time is sent to a centralized decision-making agent that utilizes the MDP framework to decide on which process to observe in the next time slot. We next describe the MDP structure and develop the process selection policy.
\subsection{Markov Decision Process}
An MDP has four components: state space, action space, state transition probabilities, reward function. In the centralized case, these components are defined as follows:
\subsubsection{MDP State} Our estimation algorithm is based on the belief vector $\vecsigma(k)$  that changes with time after each observation arrives. Therefore, we define $\vecsigma(k)\in[0,1]^N$ as the state of the MDP at time $k$. We note that the MDP state vector $\vecsigma(k)$ is different from the process state vector~$\vecs$.
\subsubsection{MDP Action} The state of MDP depends on the observation which  in turn depends on the process selected by the policy. Naturally, the action taken by the decision maker at time instant $k$ is the selected process $\calA(k)\in\{1,2,\ldots,N\}$.
\subsubsection{MDP State Transition} For our problem, the MDP state $\vecsigma(k)$ at time $k$ is a deterministic function of the previous MDP state $\vecsigma(k-1)$, the action $\calA(k)$, and the observation $\vecy_{\calA(k)}(k)$. So, the MDP state transition is modeled by \eqref{eq:sigma_update1}  and \eqref{eq:sigma_update3}.
\subsubsection{Reward Function} We seek a policy that reaches the desired level of decision accuracy with minimal stopping time $K$. Here, we capture the decision accuracy using the uncertainty associated with each process conditioned on the observations. The uncertainty associated with the $i\nth$ process can be quantified using the entropy of its posterior distribution $\begin{bmatrix}
\vecsigma_i(k) & 1-\vecsigma_i(k)
\end{bmatrix}$.  As a consequence, the instantaneous reward of the MDP is
\begin{equation}\label{eq:reward_entropy}
r_{\mathrm{entropy}}(k) = \sum_{i=1}^N H(\vecsigma_i(k-1))-H(\vecsigma_i(k)),
\end{equation}
where $H(x)=-x\log x-(1-x)\log (1-x)$ is the binary entropy function. Alternatively, we can also use the LLR of the posterior distribution of the processes to capture the decision accuracy. The alternative LLR-based instantaneous reward function is
\begin{equation}\label{eq:reward_llr}
r_{\mathrm{LLR}}(k) = \sum_{i=1}^N L(\vecsigma_i(k))-L(\vecsigma_i(k-1)),
\end{equation}
where $L(x)=x\log (x/(1-x))+(1-x)\log ((1-x)/x)$.

The differences of the entropies and LLRs quantify the reduction in uncertainty from time $k-1$ to $k$.
Hence, with these reward formulations, we encourage the RL agent to reduce the uncertainty and make a decision with high accuracy as quickly as possible. In the sequel, we use $r_{\mathrm{central}}(k)$ to denote the reward function at time $k$, which is either $r_{\mathrm{entropy}}(k)$ or $r_{\mathrm{LLR}}(k)$. Then, the long term reward can be defined as the expected discounted reward of the MDP:
$
\bar{R}(k) = \sum_{l=k}^K\gamma^{l-k} r_{\mathrm{central}}(l),$
where $\gamma\in(0,1)$ is the discount factor. The discounted reward formulation implies that a reward received $l$ time steps in the future is worth only $\gamma^l$ times what it would be worth if it were received immediately. Thus, this formulation further encourages the minimization of the stopping time.

Having defined the MDP, we next describe the actor-critic RL algorithm that solves the long-term average reward maximization problem. In the following subsection, we also describe the stopping criterion and stopping time.

\subsection{Actor-Critic Algorithm}\label{sec:deepac}
We use the deep actor-critic algorithm which is a deep learning-based RL technique to generate a sequential policy that maximizes the long-term expected discounted reward $\bar{R}(k) $ of a given MDP. The actor-critic framework maximizes the discounted reward using two neural networks: actor and critic networks.  The actor learns a stochastic policy that maps the state of the MDP to a probability vector on the set of actions. The critic learns a function that evaluates the policy of the actor and gives feedback to the actor. As a result, the two neural networks interact and adapt to each other.

The components of the actor-critic algorithm are as follows:
\subsubsection{Actor Network} The actor takes the state of the MDP $\vecsigma(k-1)\in[0,1]^N$ as its input. Its output is the probability vector $\vecmu(\vecsigma(k-1);\alpha_{\mathrm{central}})\in[0,1]^N$  over the set of processes where $\alpha_{\mathrm{central}}$ denotes the set of parameters of the actor network. The decision maker employs a stochastic process selection strategy $\calA(k)\sim \vecmu(\vecsigma(k-1);\alpha_{\mathrm{central}})$, i.e., the $i\nth$ process is selected at time $k$ with probability equal to the $i\nth$ entry $\vecmu_i(\vecsigma(k-1);\alpha_{\mathrm{central}})$ of the actor output.
\subsubsection{Reward Computation} Once the process $\calA(k)$ is selected, the decision maker receives the corresponding observation $\vecy_{\calA(k)}$, and the MDP state $\vecsigma(k-1)$ is updated to  $\vecsigma(k)$ as given by \eqref{eq:sigma_update1}. The decision maker also calculates the instantaneous reward $r_{\mathrm{central}}(k)$ using  \eqref{eq:reward_entropy} or \eqref{eq:reward_llr}, and the reward value is utilized by the critic network to provide feedback to the actor network, as discussed next.
\subsubsection{Critic Network} The critic neural network models the value function $V(\vecsigma(k))$ of the current MDP state as defined below:
\begin{equation*}
V^{\mu}(\vecsigma) = \expect{\calA(k)\sim\mu}{\bar{R}(k)\middle|\vecsigma(k)=\vecsigma}.
\end{equation*}
We note that $V^{\mu}(\vecsigma)$ is the expected average future reward when the MDP starts at state $\vecsigma$ and follows the policy $\mu(\cdot;\theta)$ thereafter. In other words, $V^{\mu}(\vecsigma)$ indicates the long-term desirability of the MDP being in state $\vecsigma$. Consequently, the input to the critic network is the posterior vector $\vecsigma\in[0,1]^N$, and the output is the learned value function $\hat{V}\lb\vecsigma;\beta_{\mathrm{central}}\rb$. Here, $\beta_{\mathrm{central}}$ is the set of parameters of the critic neural network. The scalar critique for the actor network takes the form of the temporal difference (TD) error $\delta(k;\beta_{\mathrm{central}})$ defined as follows:
\begin{equation}\label{eq:temporalerror}
\delta(k;\beta_{\mathrm{central}}) = r_{\mathrm{central}}(k)+\gamma \hat{V}(\vecsigma(k))-\hat{V}(\vecsigma(k-1)).
\end{equation} A positive TD error indicates that the probability of choosing the current action should be increased in the future, and a negative TD error suggests that the probability of choosing $\calA(k)$ should be decreased.
\subsubsection{Learning Actor Parameters} The goal of the actor is to choose a policy such that the value function is maximized which, in turn, maximizes the expected average future reward. Therefore, the actor updates its parameter set $\alpha_{\mathrm{central}}$ using the gradient descent step by moving in the direction in which the value function is maximized. The update equation for the actor parameters is given by
\begin{multline}\label{eq:policy_gradient}
\alpha_{\mathrm{central}} = \alpha_{\mathrm{central}}^-+\delta(k;\beta_{\mathrm{central}})\\\times \nabla_{\alpha_{\mathrm{central}}}[\log\mu_{\calA(k)}(\vecsigma(k-1);\alpha_{\mathrm{central}})],
\end{multline}
where $\alpha_{\mathrm{central}}^-$ is the estimate of the network obtained in the previous time instant~\cite[Ch.13]{sutton2018reinforcement}.
\subsubsection{Learning Critic Parameters} The critic chooses its parameters such that it learns the estimate $\hat{V}(\cdot)$ of the state value function $V(\cdot)$ accurately. So, the critic updates its parameter set $\beta_{\mathrm{central}}$ by minimizing the square of the TD error $\delta^2(k;\beta_{\mathrm{central}})$.
\subsubsection{Termination criterion} The actor-critic algorithm continues to collect observations until the confidence level on the decision exceeds the desired level $\upi$. We define the confidence level on $\hat{\vecs}_i$ as $\max\{\vecsigma_i(k),1-\vecsigma_i(k)\}$, and the resulting stopping criterion is as follows:
\begin{equation}\label{eq:stopping}
\underset{{i=1,2,\ldots,N}}{\min} \max\{\vecsigma_i(k),1-\vecsigma_i(k)\} > \upi.
\end{equation}

Hence, the stopping time is reached and no further observations are acquired once the stopping criterion is met. The above components completely describe the actor-critic algorithm, and we next summarize the overall algorithm and discuss its complexity.

\subsection{Overall Algorithm}\label{sec:overall}
Combining  the estimation algorithm in \Cref{sec:estimation} and the deep actor-critic method in \Cref{sec:deepac}, we obtain our centralized anomaly detection algorithm. The decision-maker collects observations using the selection policy obtained using the actor-critic algorithm until the stopping criterion given in \eqref{eq:stopping} is satisfied. After the actor-critic algorithm terminates, the decision-maker computes $\hat{\vecs}$ using \eqref{eq:estimate}. We present the pseudo-code of the overall procedure in \Cref{alg:Centralized_Training,alg:Centralized_Testing}.

\begin{algorithm}[t]
\caption{\strut Centralized actor-critic RL for anomaly detection: Training Phase}
\label{alg:Centralized_Training}
\begin{algorithmic}[1]
\REQUIRE Discount rate $\gamma\in(0,1)$, Number of time steps per episode $T$

\ENSURE Initialize $\alpha_{\mathrm{central}},\beta_{\mathrm{central}}$ randomly, $\vecsigma(0)$ with the prior probabilities on each process (can be  learned from the training data)

\FOR {Episode index $= 1,2,\ldots$}
\STATE Time index $k =1$
\REPEAT
\STATE Choose a process $\calA(k)\sim\vecmu(\vecsigma(k-1),\alpha_{\mathrm{central}})$
\STATE Receive observation $\vecy_{\calA(k)}(k)$
\STATE Compute $\vecsigma(k)$ using \eqref{eq:sigma_update1} and \eqref{eq:sigma_update3}
\STATE Compute instantaneous reward $r_{\mathrm{central}}(k)$ using \eqref{eq:reward_entropy} or \eqref{eq:reward_llr}
\STATE Update $\alpha_{\mathrm{central}}$ using \eqref{eq:policy_gradient}
\STATE Update $\beta_{\mathrm{central}}$ as the minimizer of $\delta^2(k;\beta_{\mathrm{central}})$ in \eqref{eq:temporalerror}
\STATE Increase time index $k=k+1$
\UNTIL {$k>T$}
\ENDFOR
\end{algorithmic}
\end{algorithm}

\begin{algorithm}[t]
\caption{\strut Centralized actor-critic RL for anomaly detection: Testing Phase}
\label{alg:Centralized_Testing}
\begin{algorithmic}[1]
\REQUIRE Upper threshold on confidence~$\upi$

\ENSURE $\alpha_{\mathrm{central}}$ and $\vecsigma(0)$ obtained from the training phase

\STATE Time index $k =1$
\REPEAT
\STATE Choose a process $\calA(k)\sim\vecmu(\vecsigma(k-1),\alpha_{\mathrm{central}})$
\STATE Receive observation $\vecy_{\calA(k)}(k)$
\STATE Compute $\vecsigma(k)$ using \eqref{eq:sigma_update1} and \eqref{eq:sigma_update3}
\STATE Increase time index $k=k+1$
\UNTIL {\eqref{eq:stopping} is satisfied}
\STATE Declare the estimate $\hat{\vecs}$ using \eqref{eq:estimate}
\end{algorithmic}
\end{algorithm}

The computational complexity of our algorithm is determined by the size of the neural networks, the update of the posterior belief vector given by \eqref{eq:sigma_update1} and \eqref{eq:sigma_update3}, and the reward computation given by \eqref{eq:reward_entropy} or \eqref{eq:reward_llr}. Since all of them have linear complexity in the number of processes $N$, the overall computational complexity of our algorithm is polynomial in $N$. Also, the sizes of all the variables involved in the algorithm are linear in $N$ except for the pairwise conditional probability $\bbP\ls \vecs_i\middle| \vecs_j\rs$ for $i,j=1,2,\ldots,N$. Therefore, the memory requirement of the algorithm is $\calO(N^2)$. Hence, our algorithm possesses polynomial complexity, unlike the anomaly detection algorithms in \cite{joseph2020anomaly,joseph2020anomaly2} that have exponential complexity in $N$. 

It is straightforward to extend our algorithm to the case in which the decision maker chooses $n$ processes at a time. In that case, the output layer of the actor has $\binom{N}{n}$ neurons, and we need to update $\vecsigma(k)\in[0,1]^N$ using  the conditional probabilities of the form $\bbP\ls s_{i_1}, s_{i_2},\ldots,s_{i_{n}}\middle| s_j\rs$, for $1<i_1<i_2<i_3<\ldots<N$ and $j=1,2,\ldots,N$. In this case, the overall computational  complexity of the resulting algorithm is polynomial in $N$ and the memory requirement is~$\calO(N^{n+1})$.

\section{Decentralized Decision Making and Anomaly Detection}\label{sec:decentral}
In the decentralized setting, there is no centralized decision-making agent that accumulates the observations and makes the process selection decisions. At every time instant, the sensors independently decide whether or not to observe their corresponding processes. The sensors that choose to observe the corresponding processes collect the observations. Further, depending on the underlying network topology, the sensors share their observations. In particular, the $i\nth$ sensor (i.e., the sensor corresponding to the $i\nth$ process) can receive observations from a set of neighboring sensors denoted by $\mathcal{N}_i\subseteq \{1,2,\ldots,N\}$, including the $i\nth$ sensor itself. In other words, at time $k$, the $i\nth$ sensor knows the observations corresponding to the processes indexed by $\mathcal{N}_i\cap\calA(k)$. Similarly, if the $i\nth$ sensor observes the corresponding process at time $k$ (i.e., $i\in\calA(k)$), the observation $\vecy_i(k)$ is also available at  the sensors indexed by the set $\lc j:i\in\mathcal{N}_j\rc$.

Each sensor keeps its local estimate of the marginal posterior probabilities. We denote by $\vecsigma^{(i)}(k)$ the posterior probability vector of the $i\nth$ sensor at time $k$, which is updated using $\vecy_{\mathcal{N}_i\cap\calA(k)}(k)$ via \eqref{eq:sigma_update1} and \eqref{eq:sigma_update3}.  Further,  since the sensors have potentially different posterior probability vectors, we can not directly use the stopping condition \eqref{eq:stopping}. Therefore, the $i\nth$ sensor broadcasts a message at time $k$ when the following condition is fulfilled:
\begin{equation}\label{eq:stop_local}
\max\{\vecsigma^{(i)}_i(k),1-\vecsigma^{(i)}_i(k)\} > \upi.
\end{equation}  
The sensors do not immediately stop collecting observations when \eqref{eq:stop_local} is satisfied. Instead, the sensors  continue to collect observations until they receive similar broadcast messages from all the other sensors. When the observation acquisition phase ends, each sensor $i$ declares the estimate of the corresponding process as follows:
\begin{equation}\label{eq:estimate_local}
\hat{\vecs}_i = \begin{cases}
0 & \text{ if } \vecsigma_i^{(i)}(K)\geq 1-\vecsigma_i^{(i)}(K)\\
1 & \text{ if } \vecsigma_i^{(i)}(K)< 1-\vecsigma_i^{(i)}(K),
\end{cases}
\end{equation}
where we recall that $K$ is the stopping time.

In short, in the decentralized version, the process selection algorithm runs at each sensor independently of each other. We next describe how the sensors decide whether or not to choose to observe the corresponding process. Similar to the centralized algorithm discussed in \Cref{sec:central}, we use a deep actor critic algorithm as described next.

\subsection{Decentralized Deep Actor Critic Framework and MDP}
In the decentralized algorithm, each sensor has to learn its selection policy depending on its posterior vector $\vecsigma^{(i)}(k)$. 
 Our framework consists of one actor network per sensor and a single common critic network, and we adopt the mechanism of \textit{centralized training and decentralized execution} to learn the neural networks. Specifically, we assume that all the actor networks share the same parameters and are trained together in a centralized fashion. In the testing phase, each sensor uses a separate actor network derived from the common actor network learned via centralized training. The centralized training phase assumes that all the sensors receive observations from all the other sensors. Consequently, all the sensors have the same set of observations given by $\vecy_{\calA(k)}(k)$, and they share a common posterior probability vector $\vecsigma(k)$. This assumption simplifies our training and leads to a common MDP in the centralized training phase as described next.

\subsubsection{MDP State and Action} For the centralized training phase, the MDP state and state transitions are identical to those in the centralized algorithm, i.e., we define $\vecsigma(k)\in[0,1]^N$ as the state of the MDP at time $k$, and  the MDP state transitions are modeled by \eqref{eq:sigma_update1} and \eqref{eq:sigma_update3}.  Based on the MDP state, each sensor decides whether or not to sense the corresponding process. The indices of the selected process at time $k$ denoted by $\calA(k)\subseteq\{1,2,\ldots,N\}$ represent the (joint) decisions taken by each sensor and form the MDP action.  

\subsubsection{MDP Reward} Unlike the centralized case, here $\lv\calA(k)\rv$ is not necessarily one. Since multiple processes can be observed at a given time, we introduce an associated sensing cost denoted by $\lambda>0$. Thus, at time $k$, the sensing cost of the network is $\lambda\lv\calA(k)\rv$. Here, all the sensors aim to achieve the common goal of minimum stopping time and sensing cost with the desired detection accuracy (decided by $\upi$). Therefore, we need a single reward function that promotes the common goal of the network. Consequently, we define the reward function as follows:
\begin{equation}\label{eq:reward_decentral}
r_{\mathrm{decentral}}(k) = r_{\mathrm{central}}(k)-\eta\lambda\lv\calA(k)\rv,
\end{equation}
where $\eta>0$ is the regularizer and $\lambda\lv\calA(k)\rv$ is the sensing cost of the network at time $k$. Also, we recall that $r_{\mathrm{central}}(k)$ is  either $r_{\mathrm{entropy}}(k)$ or $r_{\mathrm{LLR}}(k)$ defined in \eqref{eq:reward_entropy} and \eqref{eq:reward_llr}, respectively.
The decentralized reward function $r_{\mathrm{decentral}}(k)$ encourages the sensor network to minimize the stopping time via the first term and minimize the sensing cost via the second term.

Using the above notion of MDP, we learn the common actor and critic networks in the centralized training phase. The critic learns a function that evaluates the policy followed by the common actor and gives a common feedback to them for the joint action $\calA(k)$ of the network. The neural networks interact and adapt to each other during the centralized training phase. In the testing phase, the sensors choose the actions based on the actor network's learned policy without relying on the critic network. 

Before presenting the details of the testing and training phase, we note the similarities and differences between the MDP formulation in the centralized and decentralized settings. In the centralized setting presented in \Cref{sec:central}, the MDP captures the common posterior evolution, and the control and reward of a single decision-maker. For the decentralized case introduced and analyzed in this section, the MDP captures the evolution of the common posterior vector, the joint actions that a group of decision-makers takes, and the resulting collective reward. Further, from \eqref{eq:reward_decentral}, the difference between the reward functions of the centralized and decentralized algorithms is the additional term $\eta\lambda\lv\calA(k)\rv$ denoting the sensing cost. We note that, for the centralized algorithm, we have $\lv\calA(k)\rv=1$, and thus, the term $\eta\lambda\lv\calA(k)\rv$, if included, is independent of the selection policy and does not have any impact on the learned policy. Thus, the two reward functions in both centralized and decentralized cases are equivalent. 
 
Next, we discuss in detail the centralized training and decentralized execution phases of the decentralized decision-making framework. 

\subsection{Centralized Training}
During the centralized training phase, the common actor network takes the state of the MDP $\vecsigma(k-1)\in[0,1]^N$ as its input. The output of the common actor network is $\vecnu(\vecsigma(k-1);\alpha_{\mathrm{decentral}})\in[0,1]^N$ whose $i\nth$ entry $\vecnu_i(\vecsigma(k-1);\alpha_{\mathrm{decentral}})$ represents the probability of probing the $i\nth$ process at time $k$.  Here, $\alpha_{\mathrm{decentral}}$ denotes the set of parameters of the actor network, which is learned during the centralized training phase. 

 \begin{algorithm}[t]
\caption{\strut Decentralized actor-critic RL for anomaly detection: (Centralized) Training phase}
\label{alg:Decentralized_Training}
\begin{algorithmic}[1]
\REQUIRE Discount rate $\gamma\in(0,1)$, Number of time steps per episode $T$

\ENSURE Initialize $\alpha_{\mathrm{decentral}},\beta_{\mathrm{decentral}}$ randomly, $\vecsigma(0)$ with the prior probabilities on each process (can be  learned from the training data)

\FOR {Episode index $= 1,2,\ldots$}
\STATE Time index $k =1$
\REPEAT
\STATE $\calA(k)=\emptyset$
\FOR {process (or sensor) index $a=1,2,\ldots,N$}
\STATE Choose process $a\sim\vecnu_a(\vecsigma(k-1),\alpha_{\mathrm{decentral}})$
\IF {process $a$ is selected}
\STATE Add $a$ to $\calA(k)$ and receive observation $\vecy_{a}(k)$
\ENDIF
\ENDFOR
\STATE Compute $\vecsigma(k)$ using \eqref{eq:sigma_update1} and \eqref{eq:sigma_update3}
\STATE Compute instantaneous reward $r(k)$ using \eqref{eq:reward_decentral}
\STATE Update $\alpha_{\mathrm{central}}$ using \eqref{eq:policy_gradient_new}
\STATE Update $\beta_{\mathrm{central}}$ as the minimizer of $\delta^2(k;\beta_{\mathrm{decentral}})$  in \eqref{eq:temporalerror}
\STATE Increase time index $k=k+1$
\UNTIL {$k>T$}
\ENDFOR
\end{algorithmic}
\end{algorithm}

\begin{algorithm}[t]
\caption{\strut Decentralized local actor-critic RL for anomaly detection: Testing phase at the $i\nth$ sensor}
\label{alg:Decentralized_Testing}
\begin{algorithmic}[1]
\REQUIRE Upper threshold on confidence $\upi$

\ENSURE $\alpha_{\mathrm{decentral}}$ and $\vecsigma^{(i)}(0)$ from the training phase

\STATE Time index $k =1$
\REPEAT
\STATE Decide to observe the $i\nth$ process with probability $\vecnu_i(\vecsigma^{(i)}(k-1),\alpha_{\mathrm{decentral}})$
\IF {process $i$ is selected}
\STATE Receive observation $\vecy_{i}(k)$ and send it to sensors indexed by $\lc j:i\in\mathcal{N}_j\rc$
\ENDIF
\STATE Compute $\vecsigma(k)$ using $\vecy_{\mathcal{N}_i\cap\calA(k)}(k)$ via \eqref{eq:sigma_update1} and \eqref{eq:sigma_update3}
\IF {\eqref{eq:stop_local} is satisfied}
\STATE Broadcast a message to stop
\ENDIF
\STATE Increase time index $k=k+1$

\UNTIL {all the sensors request to stop at time $k$}
\STATE Declare the estimate $\hat{\vecs}_i$ using \eqref{eq:estimate_local}
\end{algorithmic}
\end{algorithm}

In the centralized training phase, we learn the parameters of the common actor network (denoted by $\alpha_{\mathrm{decentral}}$) and the critic network (denoted by $\beta_{\mathrm{decentral}}$). Here, the reward computation, critic network definition and learning, and termination criteria are the same as those in the centralized anomaly detection algorithm (see \Cref{sec:deepac}).  So,  the learning in the centralized training is identical to that of the centralized algorithm with a few differences in learning the parameters of the common actor network. Unlike the centralized algorithm, the output of the actor network is not a probability vector. Rather, each entry of the output is the probability of choosing the corresponding process (not the action). However, recall that the actor updates its parameter set $\alpha_{\mathrm{decentral}}$ using the gradient descent step by moving in the direction in which the value function is maximized using the probability of the action $\calA(k)$. Here, we explicitly calculate this probability as follows:
\begin{multline*}
\phi(\calA(k); \alpha_{\mathrm{decentral}}) = \displaystyle \ls\prod_{a\in\calA(k)} \vecnu_a(\vecsigma(k-1);\alpha_{\mathrm{decentral}})\rs\\
\times\ls\prod_{a\notin\calA(k)} \lb 1-\vecnu_a(\vecsigma(k-1);\alpha_{\mathrm{decentral}})\rb\rs,
\end{multline*}
where $\nu_a(\vecsigma(k-1);\alpha_{\mathrm{decentral}})$ is the probability of choosing process $a$ at time $k$, and we use the fact that each process is selected independently of others. Thus, the new update equation for the actor parameters is given by
\begin{multline}\label{eq:policy_gradient_new}
\alpha_{\mathrm{decentral}} = \alpha_{\mathrm{decentral}}^-+\delta(k;\beta_{\mathrm{decentral}})\\\times\nabla_{\alpha_{\mathrm{decentral}}}\phi(\calA(k);\alpha_{\mathrm{decentral}}),
\end{multline}
where $\alpha_{\mathrm{decentral}}^-$ is the estimate of the network obtained in the previous time instant~\cite[Ch. 13]{sutton2018reinforcement}. Further, as in the case of centralized algorithm, the critic updates its parameter set $\beta_{\mathrm{decentral}}$ by minimizing the square of the TD error $\delta^2(k;\beta_{\mathrm{decentral}})$. We describe the overall centralized procedure in 
\Cref{alg:Decentralized_Training}.

\subsection{Decentralized Execution}
For the decentralized execution during the testing phase, the actor network of the $i\nth$ sensor represents a stochastic policy that maps the posterior vector $\vecsigma^{(i)}(k)\in[0,1]^N$ to a probability $\vecnu^{(i)}(\vecsigma^{(i)}(k-1)\in[0,1]$ with which the corresponding process is selected. These actor networks are identical to the common actor except for the output layer. The reduced actor network of the $i\nth$ sensor retains only the $i\nth$ entry of the output of the common actor network, and the remaining output nodes are removed. Thus, the output of the $i\nth$ actor network is $\vecnu^{(i)}(\vecsigma(k-1);\alpha_{\mathrm{decentral}})=\vecnu_i(\vecsigma(k-1);\alpha_{\mathrm{decentral}})$ which is the probability of choosing the $i\nth$ process at time $k$. 

At every time instant $k$, the $i\nth$ sensor feeds its posterior vector $\vecsigma^{(i)}(k-1)$ as the input to its actor network and choose to observe the corresponding process with probability equal to its actor output. Depending upon the network topology (decided by $\lc\calN_i\rc_{i=1}^N$), the sensors share their observations and update the posterior vector using  \eqref{eq:sigma_update1} and \eqref{eq:sigma_update3}. The sensors continue to collect observations until all the sensors satisfy \eqref{eq:stop_local}. Thus, the overall stopping criterion of the algorithm is given by
\begin{equation}\label{eq:stopping_overall}
\underset{{i=1,2,\ldots,N}}{\min} \max\{\vecsigma_i^{(i)}(k),1-\vecsigma_i^{(i)}(k)\} > \upi,
\end{equation}
 which is similar to \eqref{eq:stopping}. After the actor-critic algorithm terminates, the decision-makers compute $\hat{\vecs}$ using \eqref{eq:estimate}. We present the pseudo-code of the overall procedure in \Cref{alg:Decentralized_Testing}.

To summarize, centralized training refers to the idea of using a single pair of common actor and critic networks in the training phase. Similarly, decentralized execution refers to the idea of using individual actor networks for independent decision-making, which are derived from the common actor network.

As a final remark on the decentralized version, we mention its relation with the centralized detection algorithms. For this, we consider the special case when all the sensors share their observations with all other sensors, i.e., for all $i\in\lc1,2,\ldots,N\rc$, we have $\mathcal{N}_i=\lc1,2,\ldots,N\rc$. In this case, the posterior vectors at all the sensors are identical, $\vecsigma^{(i)}(k)=\vecsigma(k)$, for all values of $i\in\lc1,2,\ldots,N\rc$ and $k>0$. Mathematically, this system is equivalent to a centralized anomaly detection algorithm that uses the common actor learned in the centralized training phase to select the processes. This centralized algorithm chooses to observe the $i\nth$ process with probability given by the corresponding entry of the actor output $\vecnu_i(\vecsigma(k-1);\alpha_{\mathrm{decentral}})$. Nonetheless, we note that this algorithm (which can potentially have $\calA(k)>1$)  is different from the centralized algorithm discussed in \Cref{sec:central} (which restricts $\calA(k)=1$). 
Moreover, it is important to note that our decentralized framework can address a spectrum of scenarios ranging from complete information exchange (as discussed above) to no information exchange (in which each sensor makes its decision based only on local observations, and hence, $\mathcal{N}_i = \{i\}$ for all $i$). Indeed, these two cases (referred to as shared detection algorithm and local detection algorithm, respectively) are analyzed in the numerical results in \Cref{sec:decentral_sim}.

\section{Simulation Results}\label{sec:simulations}
In this section, we empirically study the detection performance of our algorithm. We use three metrics for the performance evaluation: accuracy (the fraction of times the algorithm correctly identifies all the anomalous processes), stopping time ($K$), and number of observations per unit time ($=\frac{1}{K}\sum_{k=1}^K\lv\calA(k)\rv$).

Our simulation setup is as follows. We consider five processes $N=5$ and assume that the probability of each process being normal is $q=0.8$. Here, the first and second processes (or equivalently $\vecs_1$ and $\vecs_2$) are statistically dependent, and the third and fourth processes ($\vecs_3$ and $\vecs_4$) are also statistically dependent. These pairs of processes are independent of each other and independent of the fifth process ($\vecs_5$). The dependence is captured using the correlation coefficient $\rho\in[0,1]$ that is common to both process pairs:
 \begin{align*}
 \bbP\ls \vecs_1=\vecs_2=0\rs &= \bbP\ls \vecs_3=\vecs_4=0\rs = q^2+\rho q(1-q)\\
\bbP\ls \vecs_1=\vecs_2=1\rs &= \bbP\ls \vecs_3=\vecs_4=1\rs = (1-q)^2+\rho q(1-q)\\
 \bbP\ls \vecs_1\neq \vecs_2\rs &=  \bbP\ls \vecs_3\neq \vecs_4\rs =(1-\rho) q(1-q).
 \end{align*}
 Also, we assume that the flipping probability is $p=0.2$.

 Before we present the simulation results, we note that for the above setting, the pairwise probabilities required for the posterior updates in \eqref{eq:sigma_update3} of the estimation can be computed as follows. Here, there are three independent groups of  process $\lb \{1,2\}, \{3,4\}, \{5\}\rb$. If $i$ and $j$ are independent processes, for any $s,s'\in\{0,1\}$, we have
 \begin{equation*}
 \bbP\ls \vecs_i =s'\middle| \vecs_j=s\rs = \bbP\ls \vecs_{i} =s'\rs = \begin{cases}
 q &\text{if } s'=0\\
 1-q &\text{if } s'=1.
 \end{cases}
 \end{equation*}
Similarly, if $i$ and $j$ are dependent, the pairwise probabilities are given by
\begin{align*}
\bbP\ls \vecs_{i} =s'\middle| \vecs_j=s\rs &= \frac{\bbP\ls \vecs_{i} =s', \vecs_i=s\rs}{\bbP\ls \vecs_i=s\rs} \\
&= \begin{cases}
q+\rho (1-q)& s'=s=0\\
(1-q)+\rho q& s'=s=1\\
(1-\rho) q & s'=0,s=1\\
(1-\rho) (1-q) & s'=1,s=0.
\end{cases}
\end{align*}
Also, the prior distribution $\vecsigma(0) = q\one$.

 In the following subsections, we present the numerical results for the centralized and decentralized algorithms. We note that the scalable centralized algorithm chooses only one process per unit time whereas the scalable decentralized algorithm can potentially choose multiple processes per unit time. So, there is no direct comparison between the two presented algorithms. For a fair comparison, we compare each of the algorithms with the competing algorithms that also follow similar policies.

\begin{figure*}[hptb]
\begin{center}

\begin{subfigure}{5.5cm}
\begin{center}
\hspace{-0.3cm}\includegraphics[height= 5.3cm]{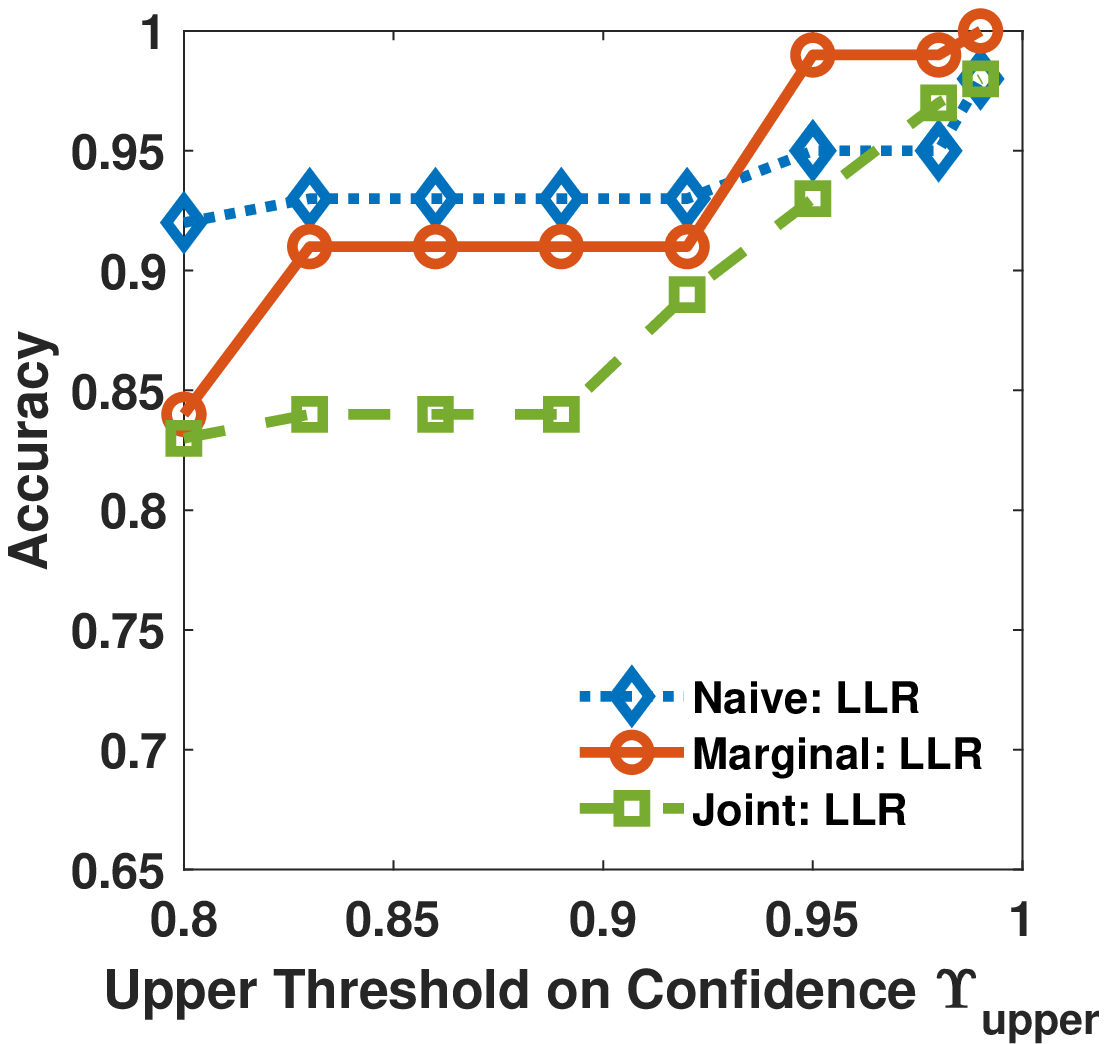}

\vspace{0.5cm}
\includegraphics[height= 5.3cm]{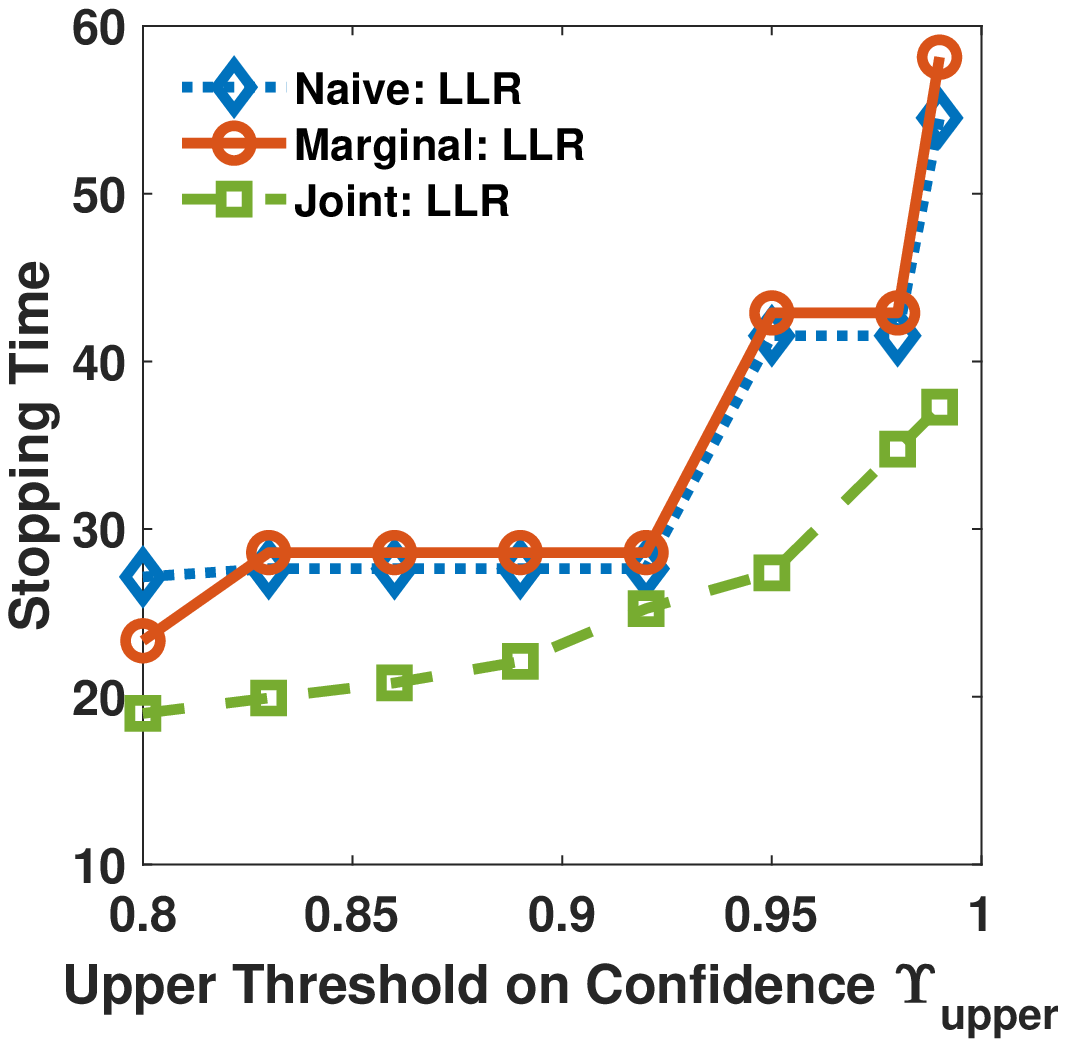}
\caption{$\rho = 0$}

\end{center}
\end{subfigure}
\begin{subfigure}{5.5cm}
\begin{center}
\hspace{-0.3cm}\includegraphics[height= 5.3cm]{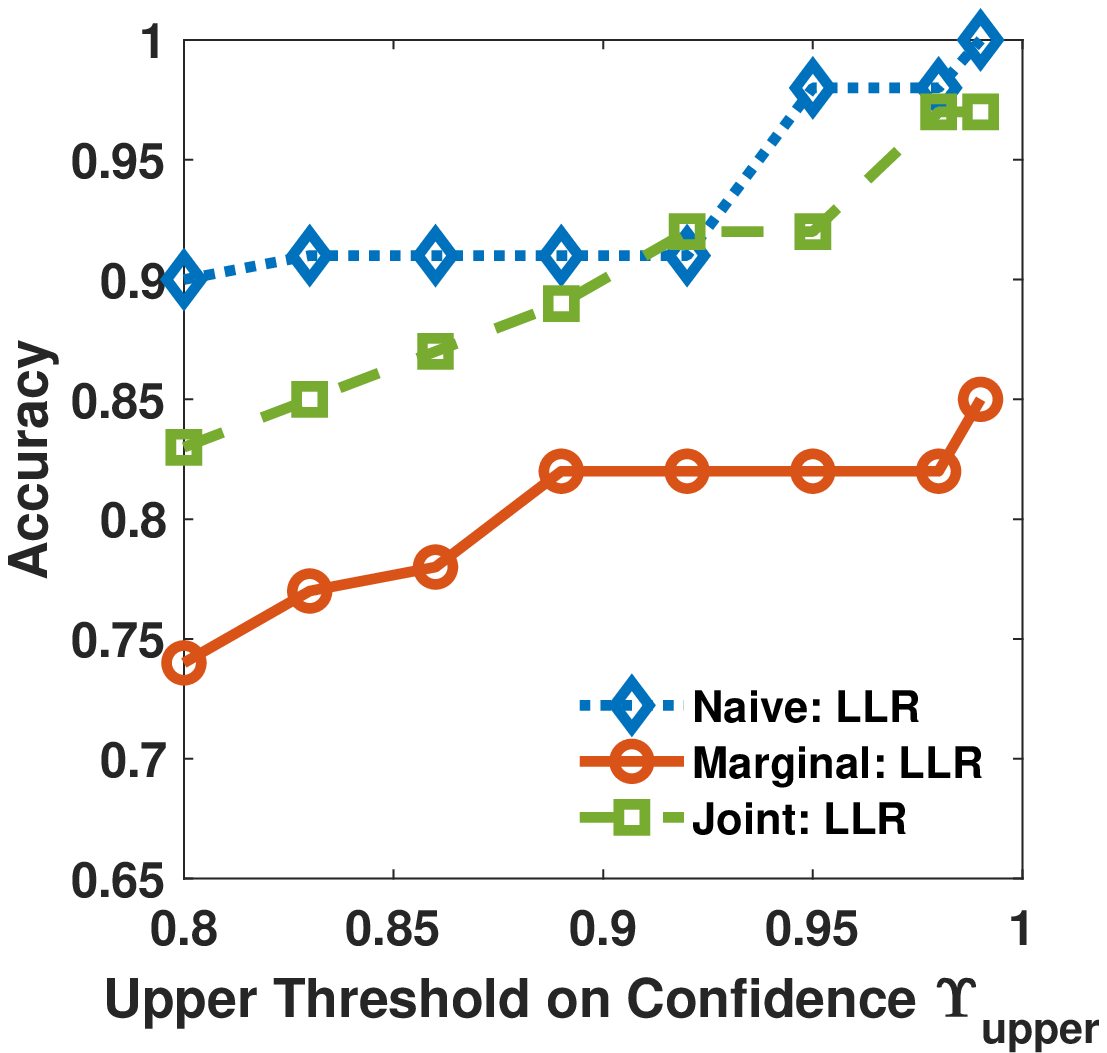}

\vspace{0.5cm}
\includegraphics[height= 5.3cm]{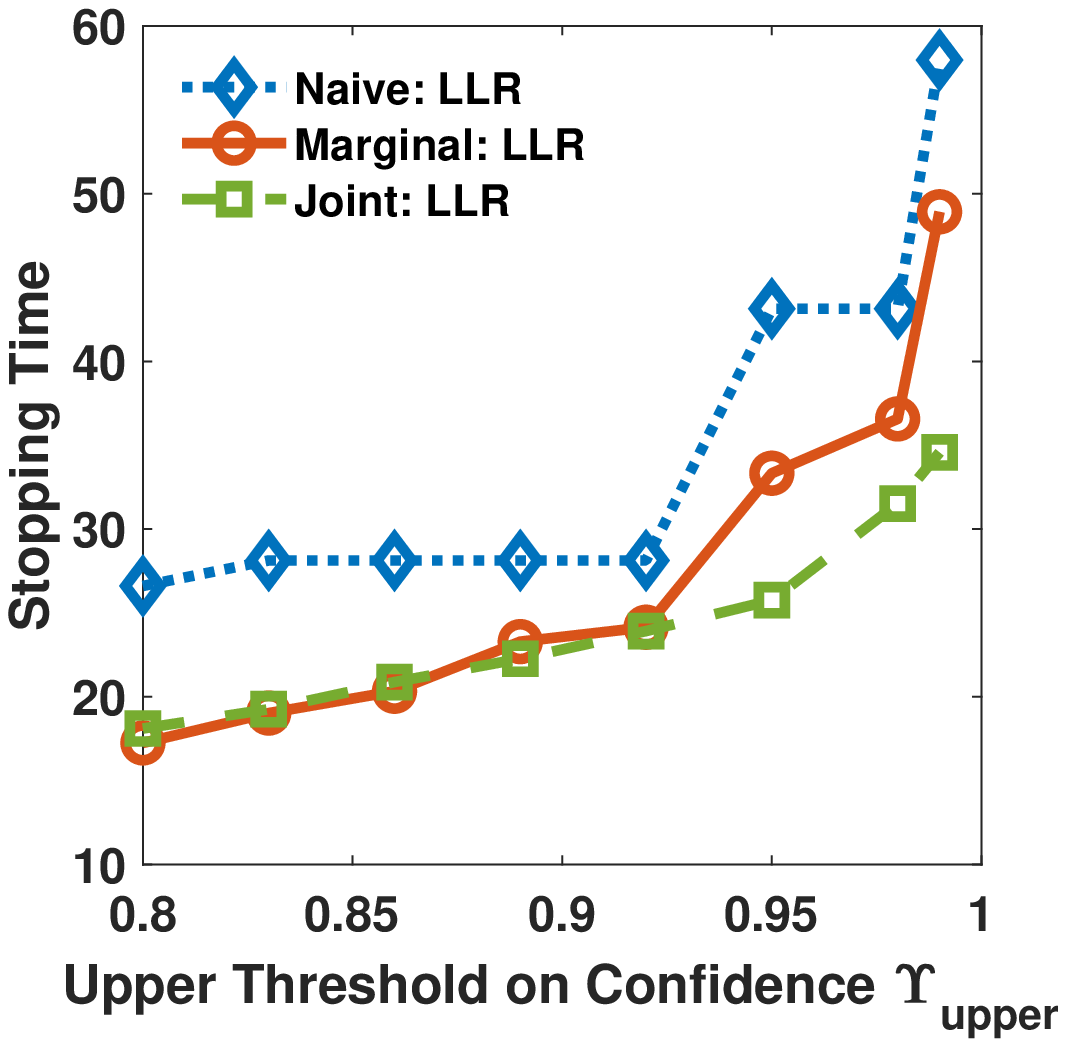}
\caption{$\rho = 0.6$}
\end{center}
\end{subfigure}
\begin{subfigure}{5.5cm}
\begin{center}
\hspace{-0.3cm}\includegraphics[height= 5.3cm]{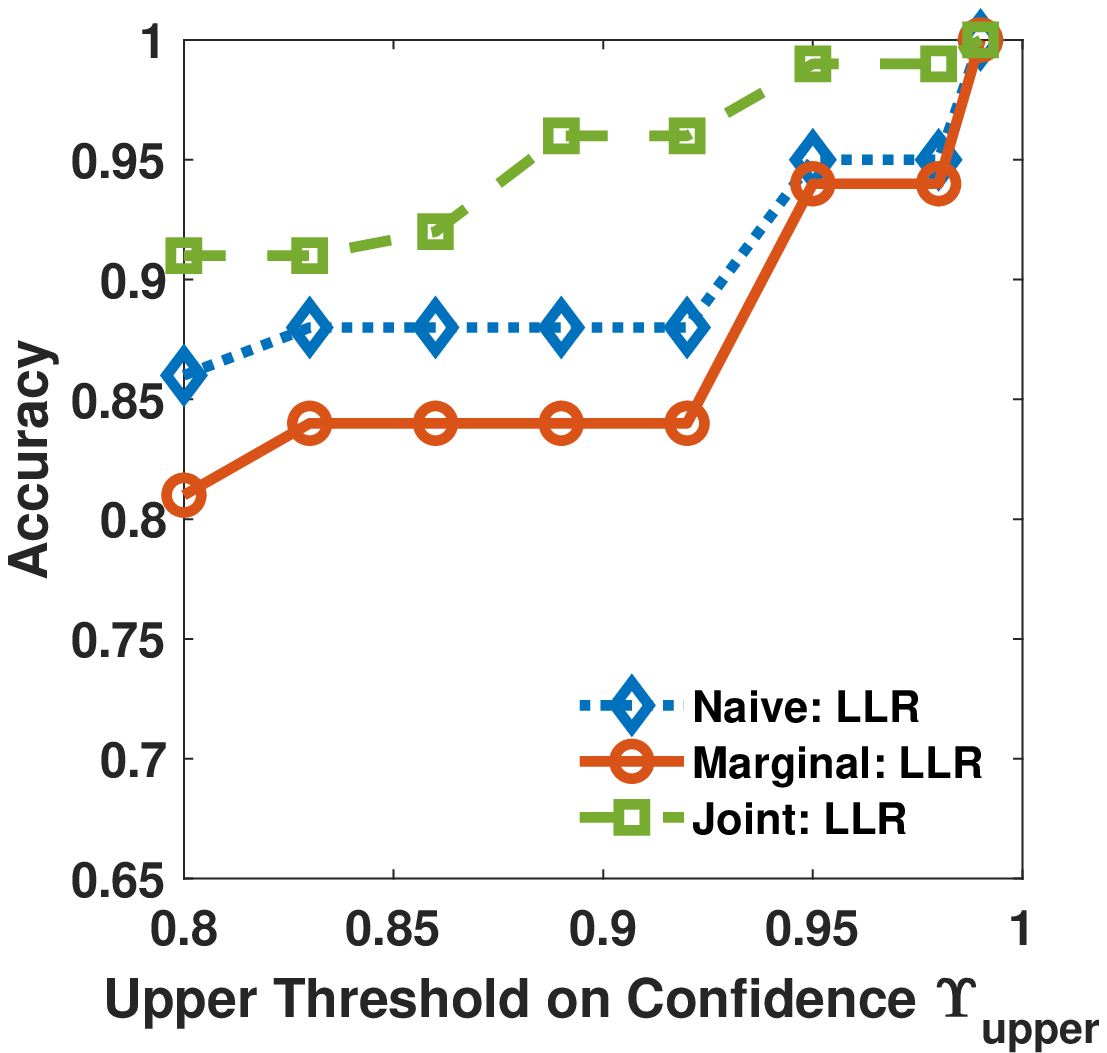}

\vspace{0.5cm}
\includegraphics[height= 5.3cm]{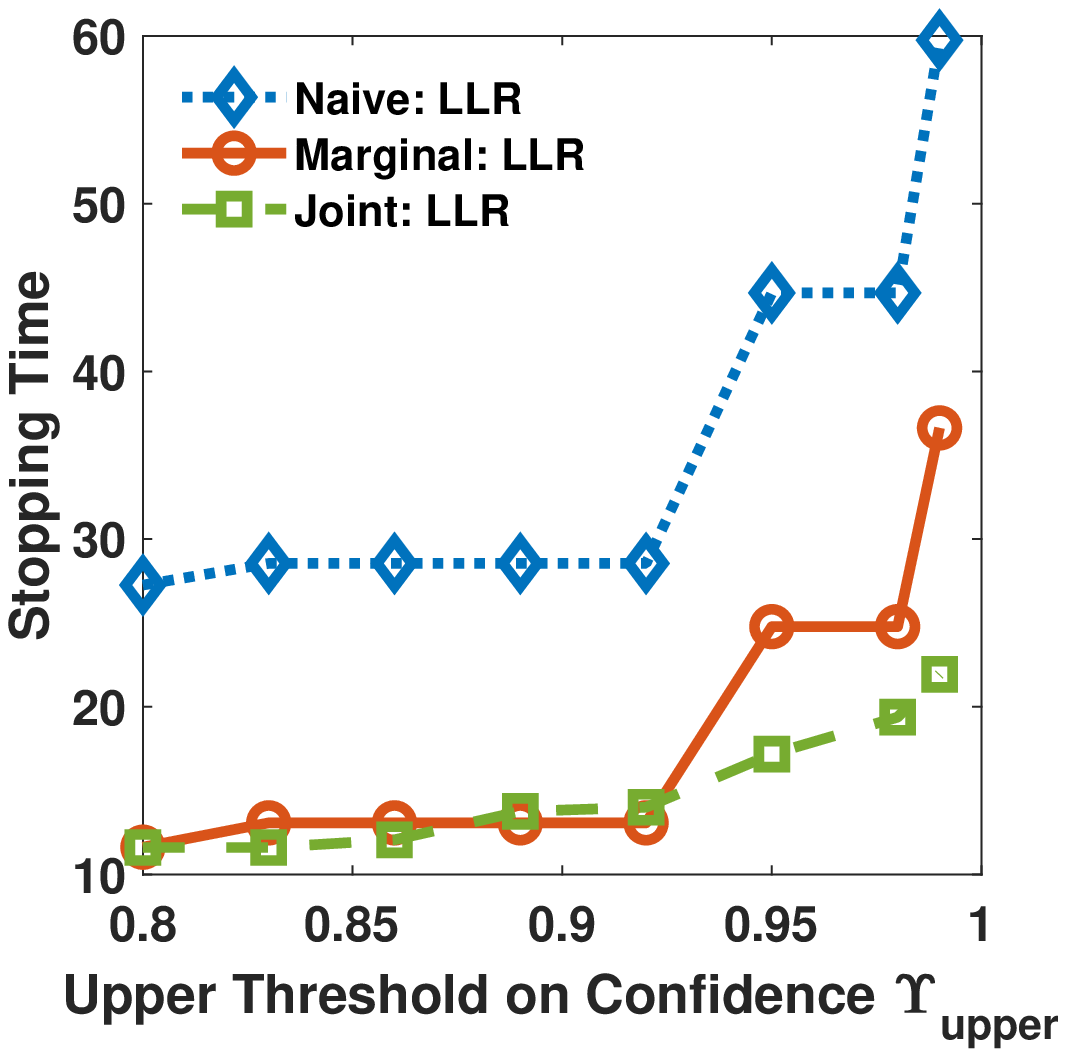}
\caption{$\rho = 1$}

\end{center}
\end{subfigure}

\end{center}

\caption{Performances of the centralized deep actor-critic algorithms using LLR-based reward as a function of  $\upi$ for different values of correlation coefficient $\rho$.}
\label{fig:ThresVs}
\end{figure*}

\begin{figure*}[hpbt]
\vspace{-0.18cm}
\begin{center}

\begin{subfigure}[b]{5.5cm}
\includegraphics[height= 5.3cm]{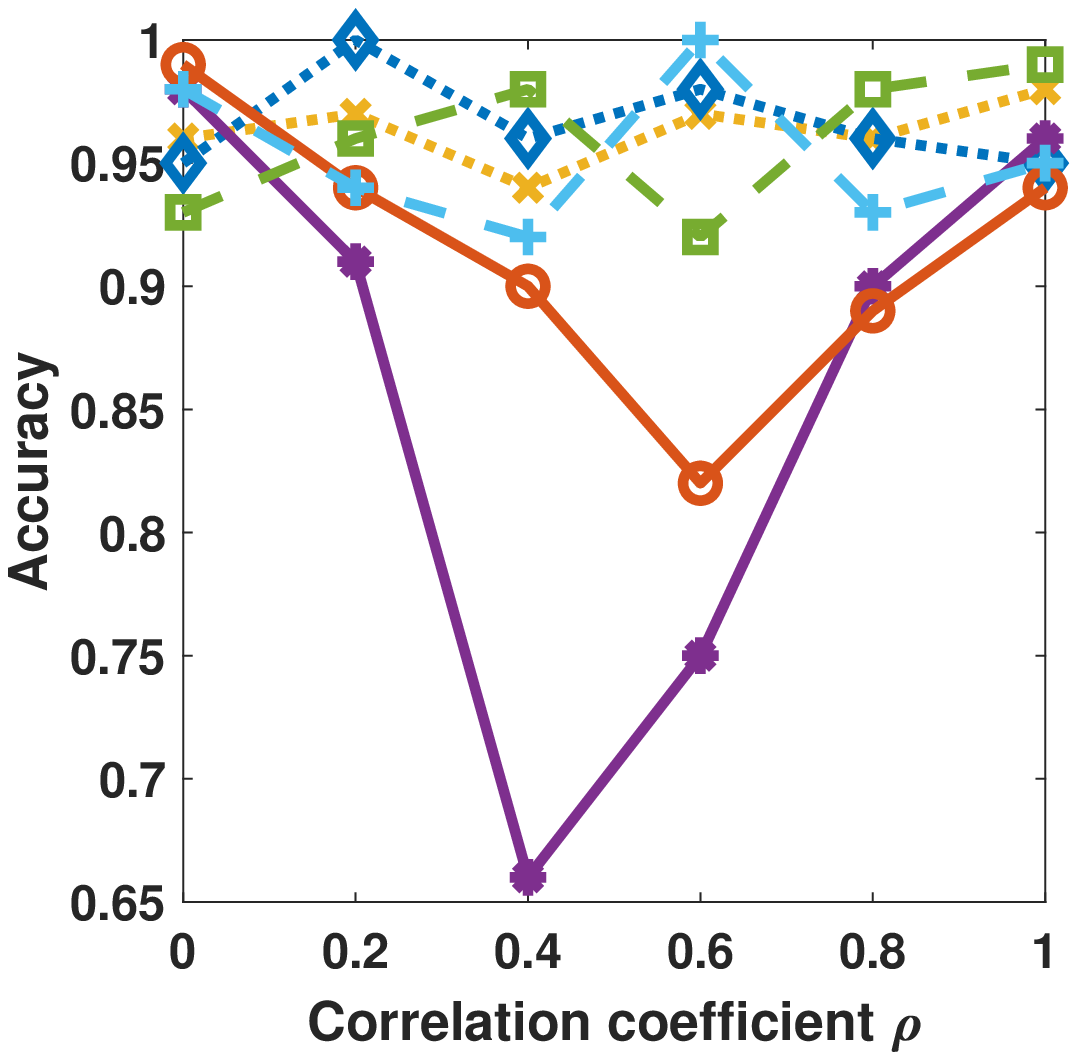}
\end{subfigure}
\begin{subfigure}[b]{5.5cm}
\includegraphics[height= 5.3cm]{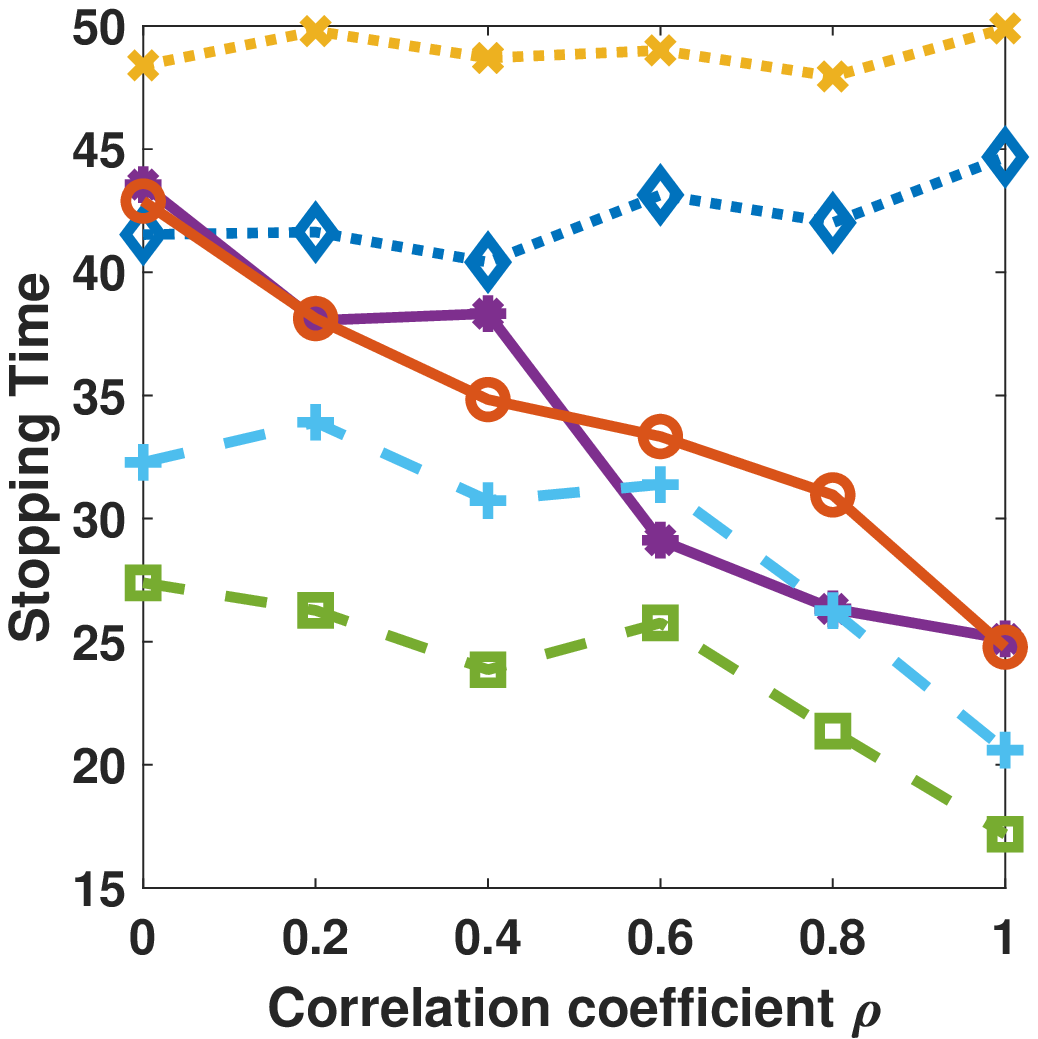}
\end{subfigure}
\begin{subfigure}[b]{4cm}
\includegraphics[width = 3.8cm]{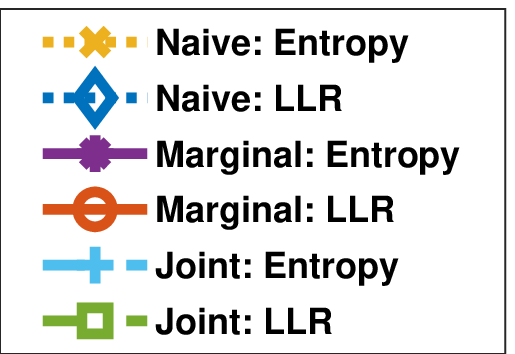}
\end{subfigure}

\end{center}
\caption{Performances of different centralized deep actor-critic algorithms with $\rho$ as a function of  $\upi = 0.95$.}
\label{fig:RhoVs}
\end{figure*}

\subsection{Centralized Algorithm}
The architecture and parameters of our centralized algorithm are as follows. We implement the actor and critic neural networks with three layers and the ReLU activation function between consecutive layers. The output layer of the actor layer is normalized to ensure that $\mu(\cdot)$ is a probability vector over the set of processes. The parameters of the neural networks are updated using Adam Optimizer, and we set the learning rates of the actor and the critic as $5\times 10^{-4}$, and $5\times 10^{-3}$, respectively. Also, we set the discount factor as $\gamma = 0.9$.

We compare the performance of our algorithm (labeled as \texttt{Marginal} due to the use of marginal posterior probabilities) with two other deep actor-critic-based schemes that also choose one process per unit time.
\begin{itemize}
\item \emph{Joint probability mass function (pmf)-based scheme (labeled as \texttt{Joint}):} This algorithm refers to the state-of-the-art method for anomaly detection problem presented in \cite{joseph2020anomaly}. The algorithm is based on the joint posterior probabilities of all the entries of $\vecs\in[0,1]^N$. Since $\vecs$ can take $2^N$ possible values, the complexity of this algorithm is exponential in $N$. However, the joint probabilities help the algorithm to learn all possible statistical dependencies among the process.
\item\emph{Naive marginal pmf-based scheme (labeled as \texttt{Naive}):} We also consider a naive method that relies on the marginal posterior probabilities $\vecsigma\in[0,1]^N$. This algorithm is identical to our algorithm except that at every time instant, this method only updates the entry $\vecsigma_{\calA(k)}(k)$ of $\vecsigma(k)$ corresponding to the selected process $\calA(k)$. In other words, this method ignores the possible statistical dependence of the observation $\vecy_{\calA(k)}(k)$ on the processes other than $\calA(k)$. Hence, the computational complexity of this algorithm is also $\calO(N)$. We note that, unlike our algorithm, this algorithm does not use any approximation, and its updates are always exact.

\end{itemize}

We implement the three algorithms using both entropy and LLR-based reward functions (see \eqref{eq:reward_entropy} and \eqref{eq:reward_llr} for the marginal pmf-based algorithms and \cite{joseph2020anomaly} for the joint pmf-based algorithm). Our centralized algorithm is a compromise between the above two algorithms and relies on marginal probabilities $\vecsigma(k)$ while accounting for the possible statistical dependence among the processes.



We note that the number of observations per unit time for the centralized algorithm is fixed to be one, and therefore, we use the accuracy and stopping time as the performance metrics. Our results are summarized in \Cref{fig:ThresVs,fig:RhoVs}  and \Cref{tab:runtime}, and the key inferences from them are discussed next.

In \Cref{fig:ThresVs}, we plot the accuracy (in the first row) and the stopping time (in the second row) as a function of the upper threshold $\upi$ for confidence.\footnote{To make the figures clearer, we omitted the curves corresponding to the entropy based reward function. The interested readers can refer to~\cite{joseph2020scalable} for similar figures for the entropy based reward function. The performances of our algorithm based on the two reward functions have similar trends under all settings. However, the LLR-based scheme slightly outperforms the entropy-based scheme in most cases.} In different columns of  \Cref{fig:ThresVs}, the correlation coefficient $\rho$ differs. The accuracy and the stopping time of all the algorithms increase with $\upi$. This trend is expected due to the fact that as $\upi$ increases, the decision-maker requires more observations to satisfy the higher desired confidence level. In \Cref{fig:RhoVs}, we plot the accuracy and stopping time curves as a function of the $\rho$. Comparing the two reward functions, we infer that LLR-based scheme slightly outperforms the entropy-based scheme.  

From \Cref{fig:RhoVs} and along with the results in \Cref{fig:ThresVs}, we next look at the dependence of the algorithm performance on $\rho$. We notice that the accuracy of our algorithm is comparable to the other two algorithms when $\rho=0$ and $\rho=1$. The accuracy degrades as $\rho$ is close to 0.5. This behavior is because our algorithm uses approximate marginal probabilities to compute the confidence level whereas the other two algorithms use exact values. This approximation in \eqref{eq:approx} is exact when $\rho=0$ and $\rho=1$. As $\rho$ approaches $0.5$, the approximation error increases, and the accuracy decreases. Also, the stopping times of the three algorithms are similar when $\rho=0$. This behavior is because when $\rho=0$, all the processes are independent, and the updates of our algorithm are exact. The naive marginal pmf-based algorithm also offers good performance when $\rho=0$ as there is no underlying statistical dependence among the processes. Further, the stopping times of our algorithm and the joint pmf-based algorithm improve with $\rho$. As $\rho$ increases, the processes become more correlated, and therefore, an observation corresponding to one process has more information about the other correlated processes. However, the naive marginal pmf-based algorithm ignores this correlation and handles the observations corresponding to the different processes independently. Therefore, the stopping time is insensitive to $\rho$. Consequently, the difference between the stopping times of the naive marginal pmf-based algorithm and the other two algorithms increases as $\rho$ increases. 

\begin{table}[hptb]
\caption{Comparison of average (testing phase) runtime (in ms) per process selection for different centralized schemes}
\begin{center}
\small
\begin{tabular}{|*{4}{c|}}
\hline
\bf  \begin{tabular} {c}Reward \\ function \end{tabular}& \bf \begin{tabular}{c}
Naive \\
marginal
\end{tabular}  & \bf \begin{tabular}{c}
Our \\ marginal
\end{tabular} & \bf \begin{tabular}{c}
Joint
\end{tabular} \\
\hline
Entropy& 0.46  & 0.49 &   0.68 \\
\hline
LLR & 0.48 &0.51& 0.69 \\
\hline
\end{tabular}

\end{center}
\label{tab:runtime}
\end{table}

The run times of algorithms given in \Cref{tab:runtime} demonstrate that the joint pmf-based algorithm is computationally heavier (40\% higher) compared to the other two algorithms. This observation is in agreement with our complexity analysis in \Cref{sec:overall}. We also recall that the difference between the runtimes of the joint pmf-based algorithm and our algorithm grows with $N$.

Thus, we conclude that our algorithm combines the best of two worlds by benefiting from the statistical dependence among the processes (similar to the joint pmf-based algorithm) and offering low-complexity (similar to the naive marginal pmf-based algorithm).

\begin{figure*}[hptb]
\begin{subfigure}{\textwidth}
\begin{center}
\includegraphics[height= 5.4cm]{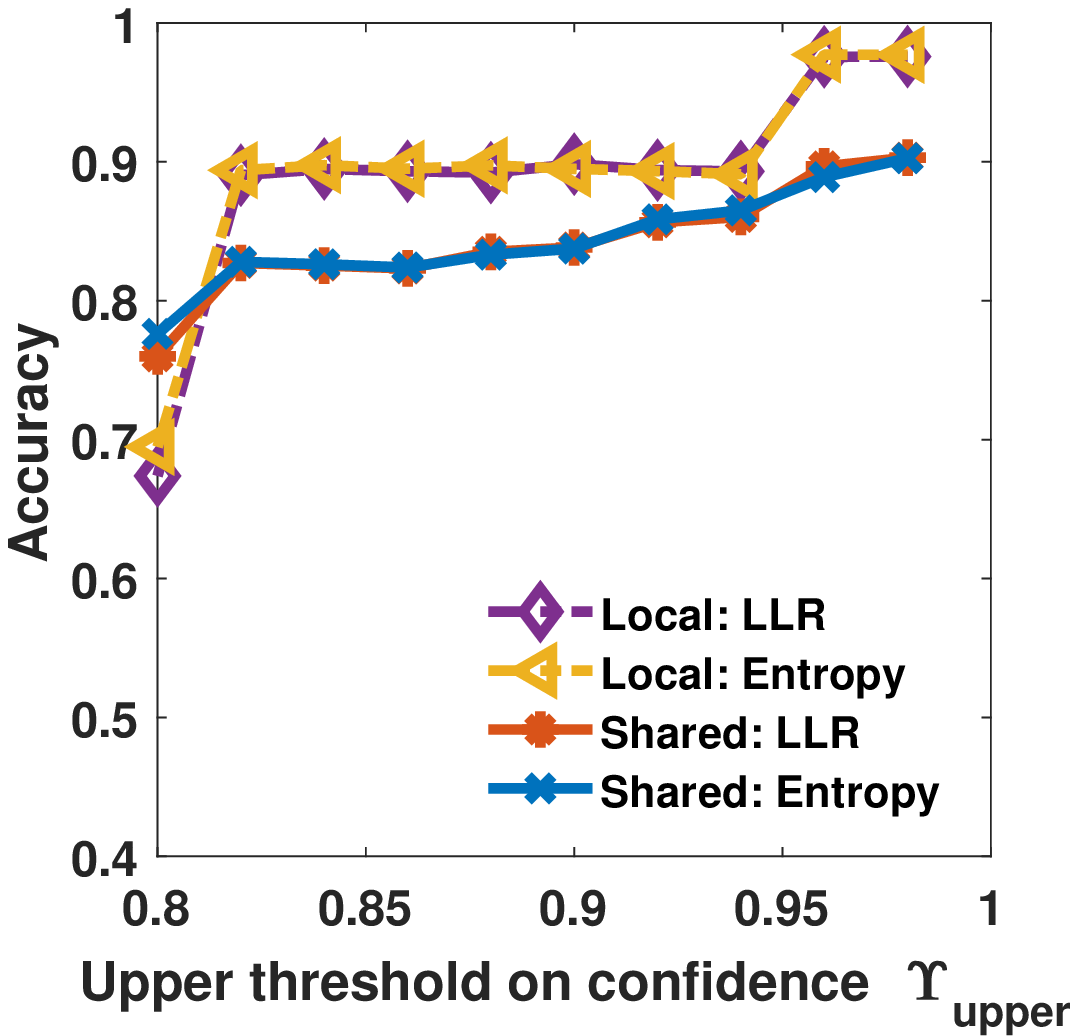}
\includegraphics[height= 5.4cm]{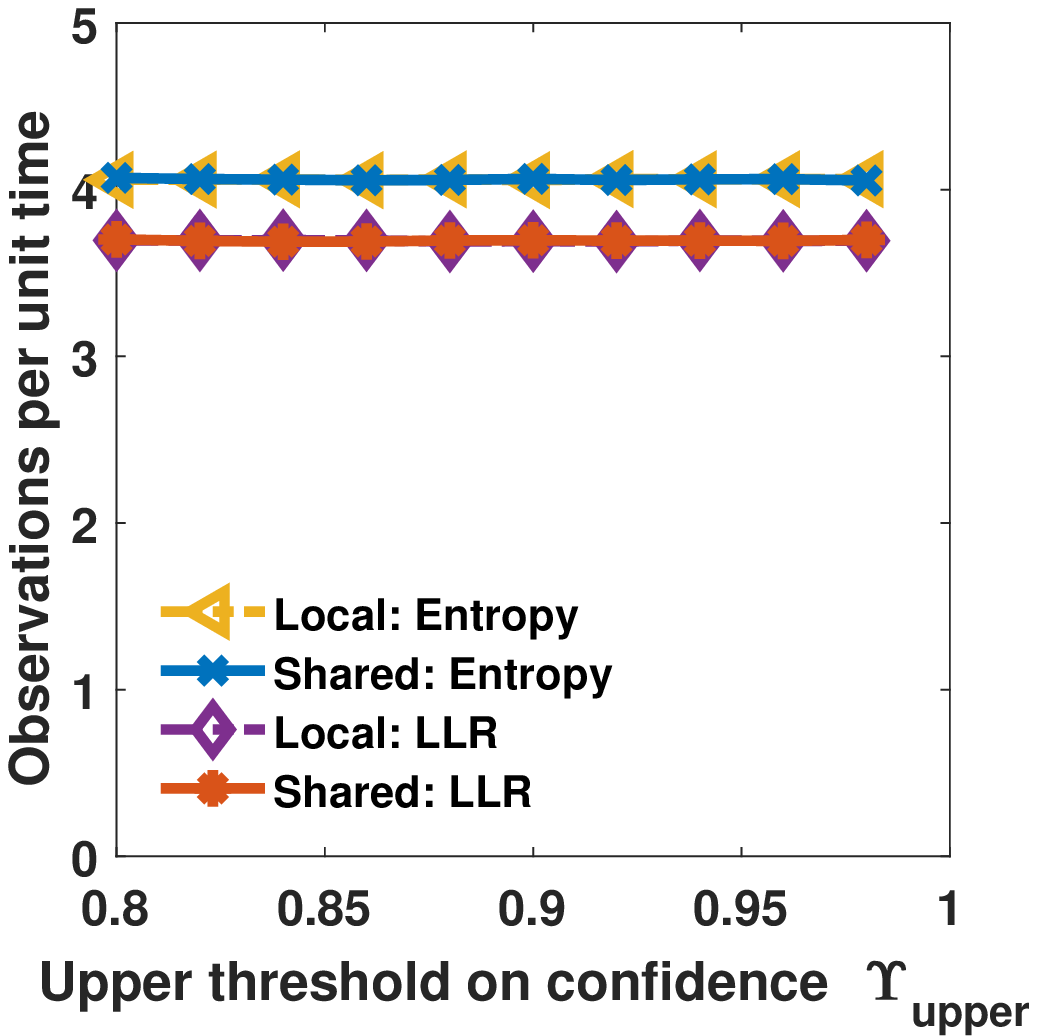}
\hspace{-0.5cm}
\includegraphics[height= 5.4cm]{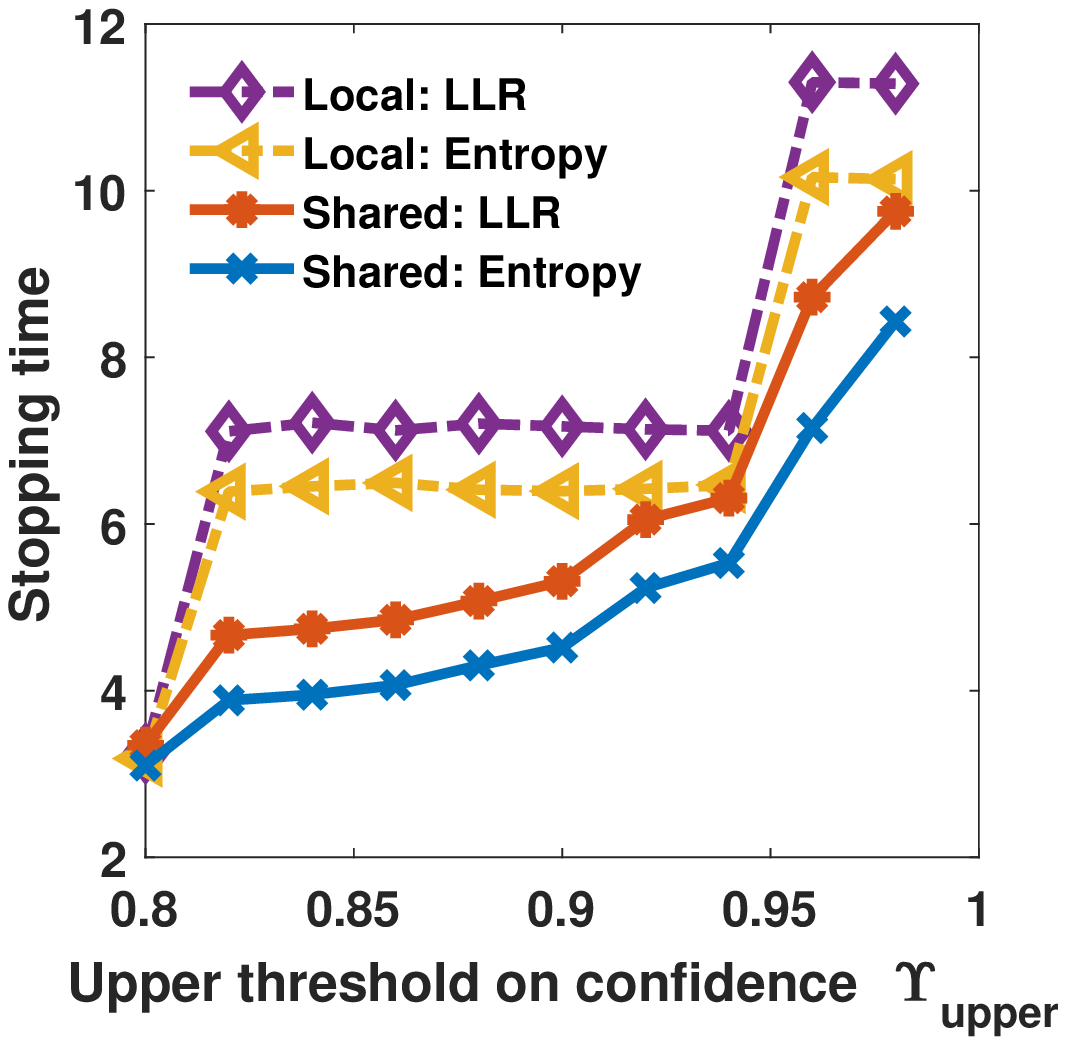}

\caption{Varying $\upi$ when $\rho=0.8$ and $\lambda=5$.}
\end{center}
\end{subfigure}

\vspace{0.7cm}

\begin{subfigure}{\textwidth}
\begin{center}
\includegraphics[height= 5.4cm]{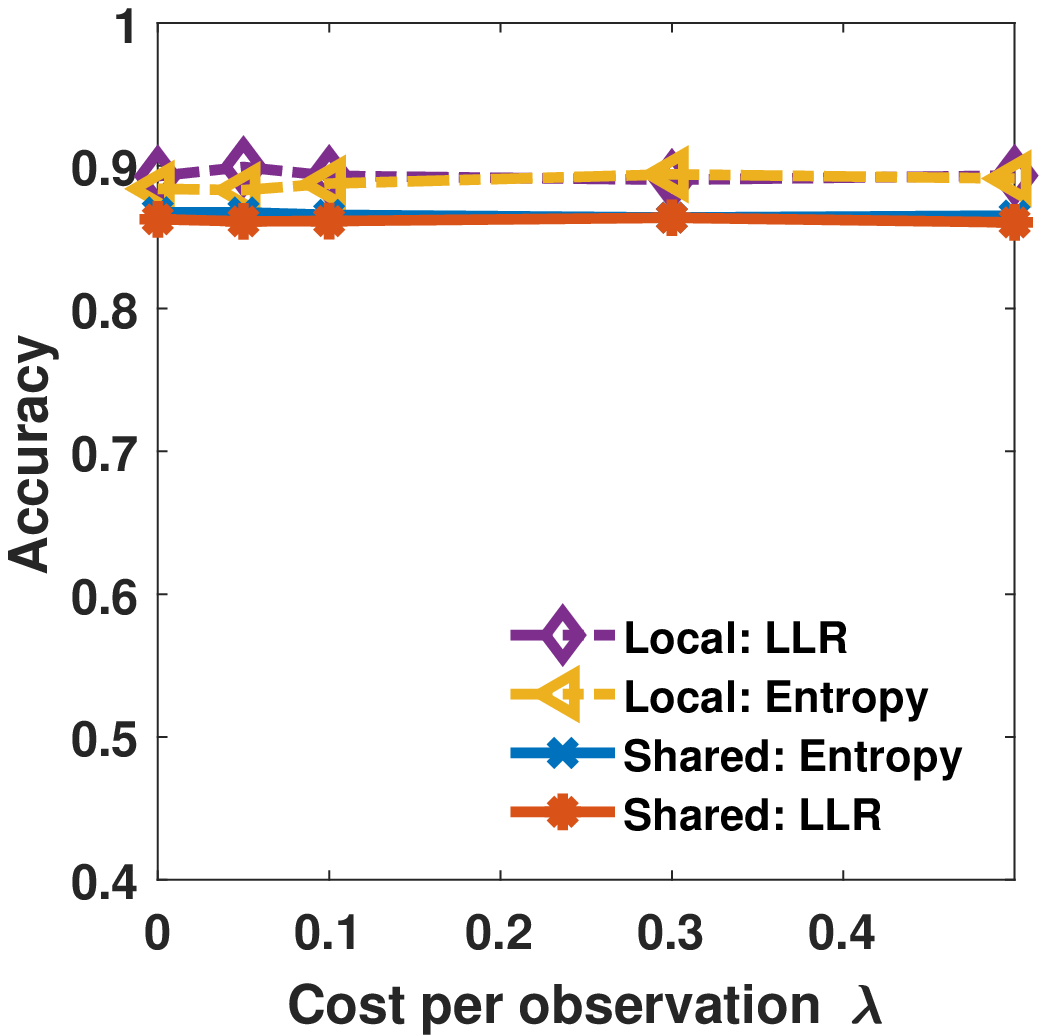}
\includegraphics[height= 5.4cm]{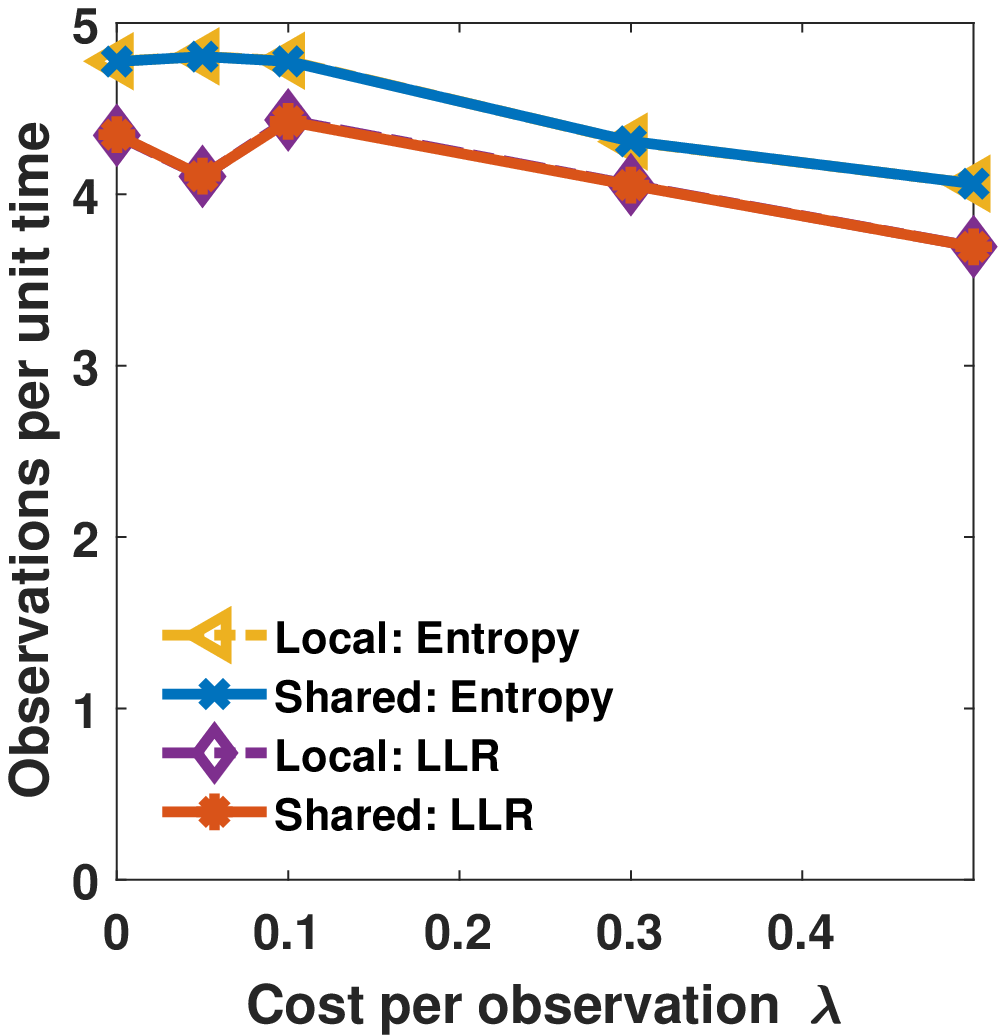}
\includegraphics[height= 5.4cm]{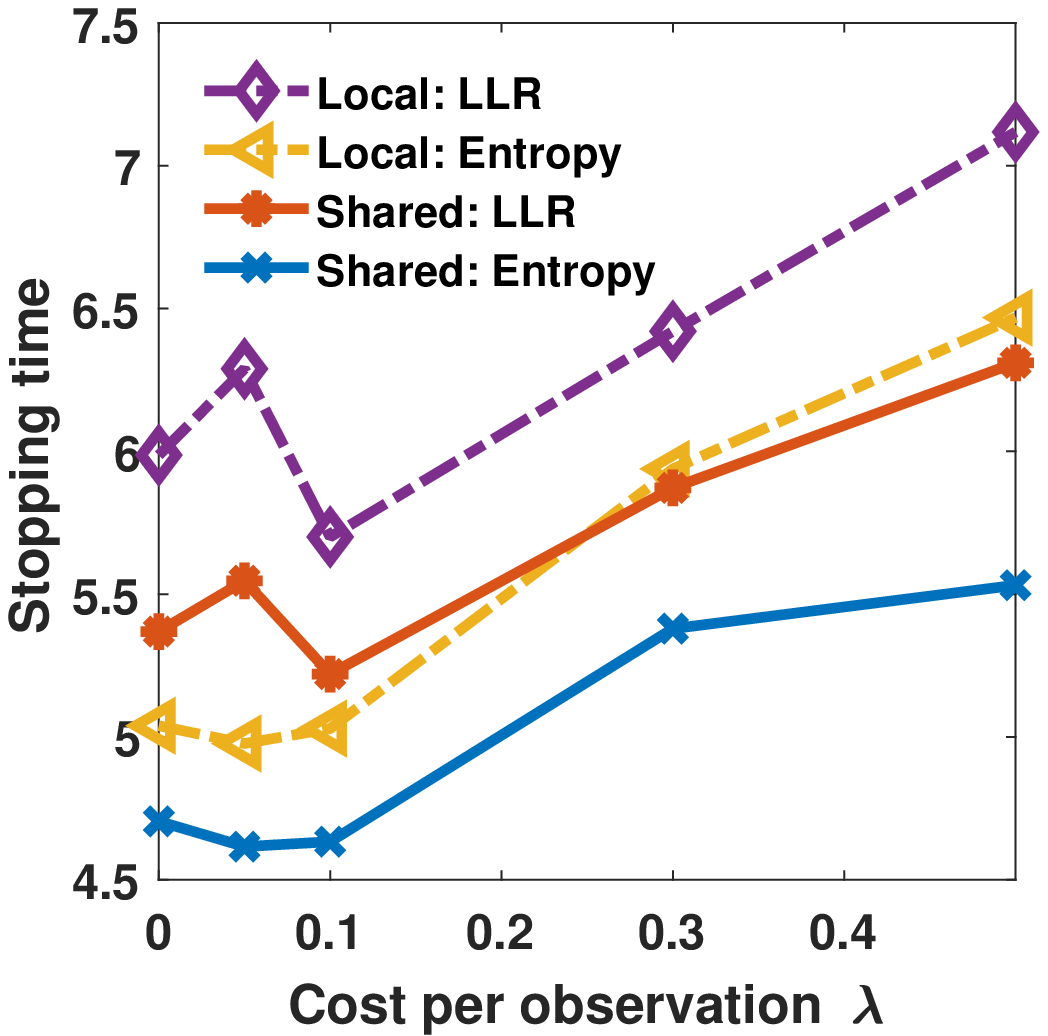}

\caption{Varying $\lambda$ when $\upi=0.95$ and $\rho=0.8$.}
\end{center}
\end{subfigure}

\vspace{0.7cm}

\begin{subfigure}{\textwidth}
\begin{center}
\includegraphics[height= 5.4cm]{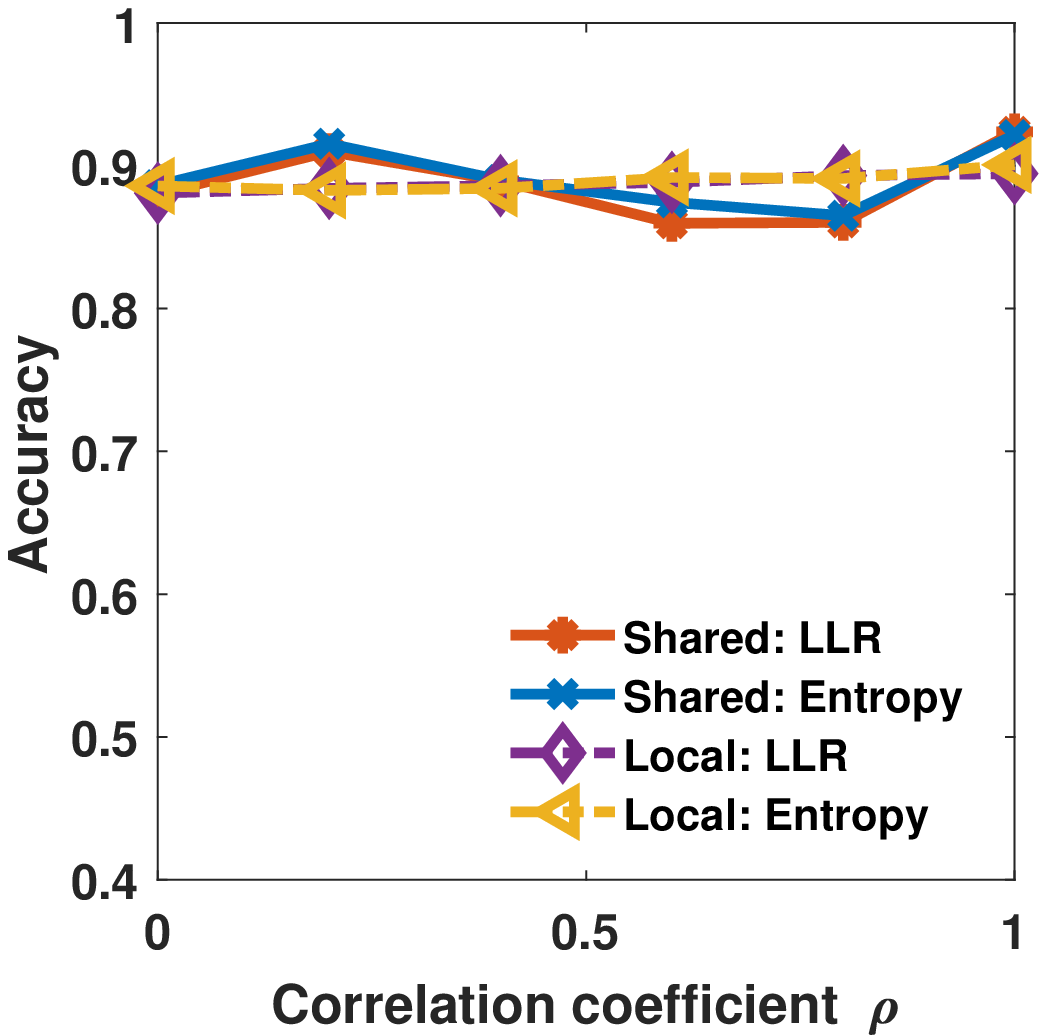}
\includegraphics[height= 5.4cm]{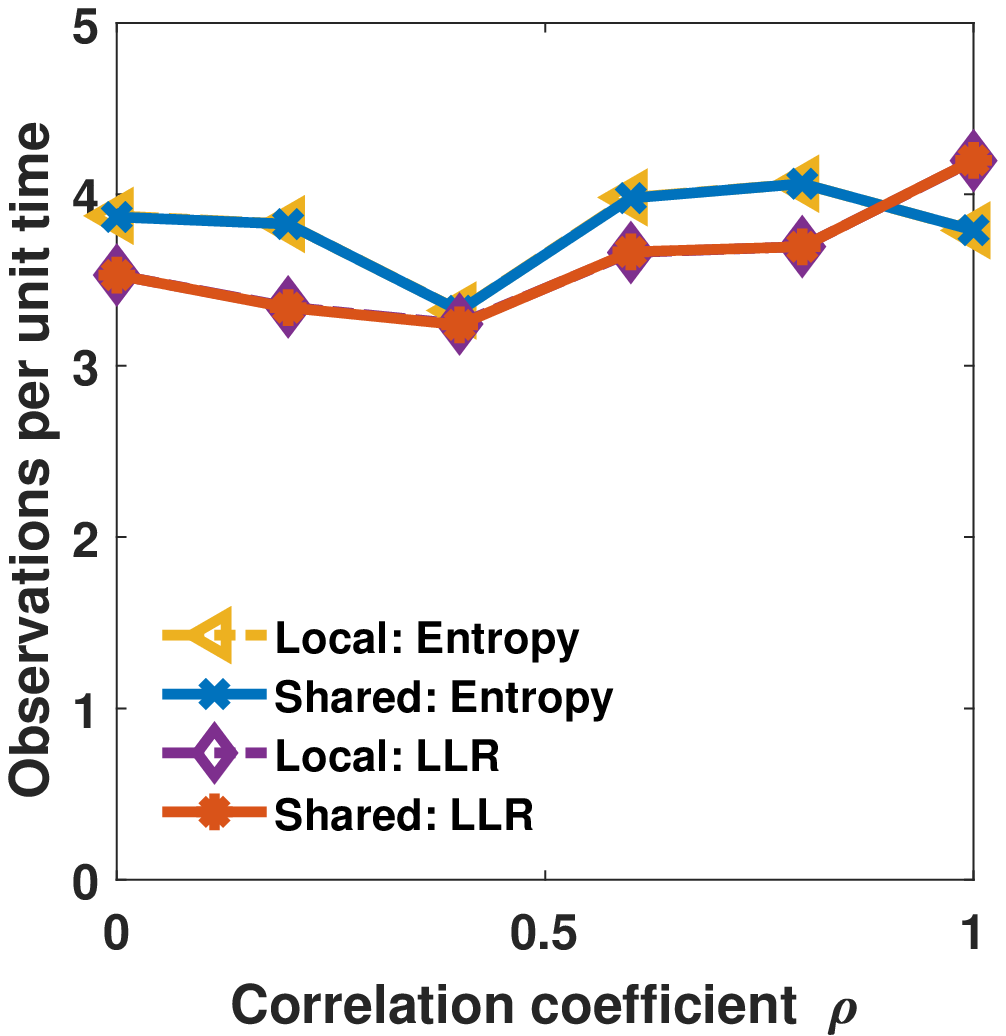}
\includegraphics[height= 5.4cm]{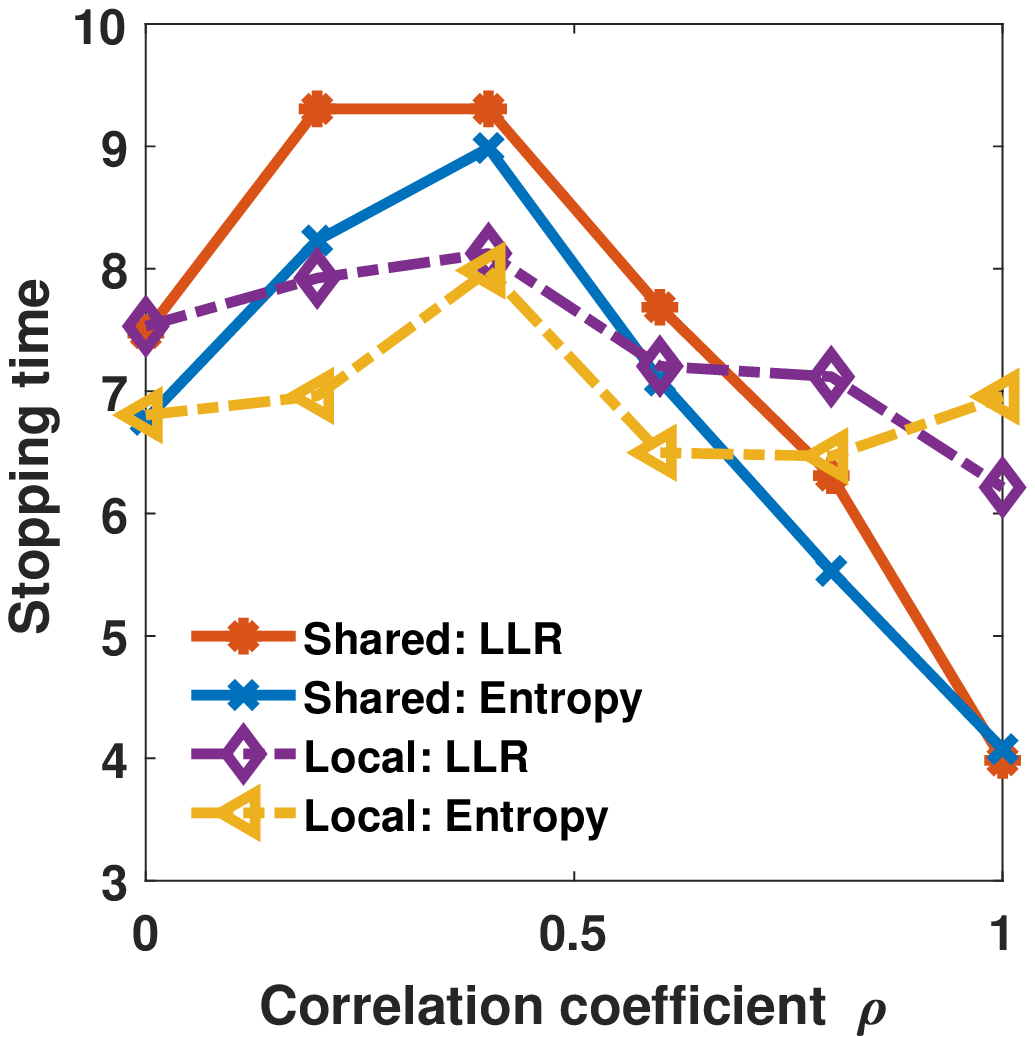}

\caption{Varying $\rho$ when $\upi=0.95$ and $\lambda=5$.}

\end{center}
\end{subfigure}

\caption{Performances of the shared and local detection algorithms as a function of the upper threshold on confidence $\upi$, sensing cost per observation $\lambda$ and correlation coefficient $\rho$.}
\label{fig:Decentralized}
\end{figure*}

\begin{figure*}[hptb]
\begin{center}
\includegraphics[height= 5.4cm]{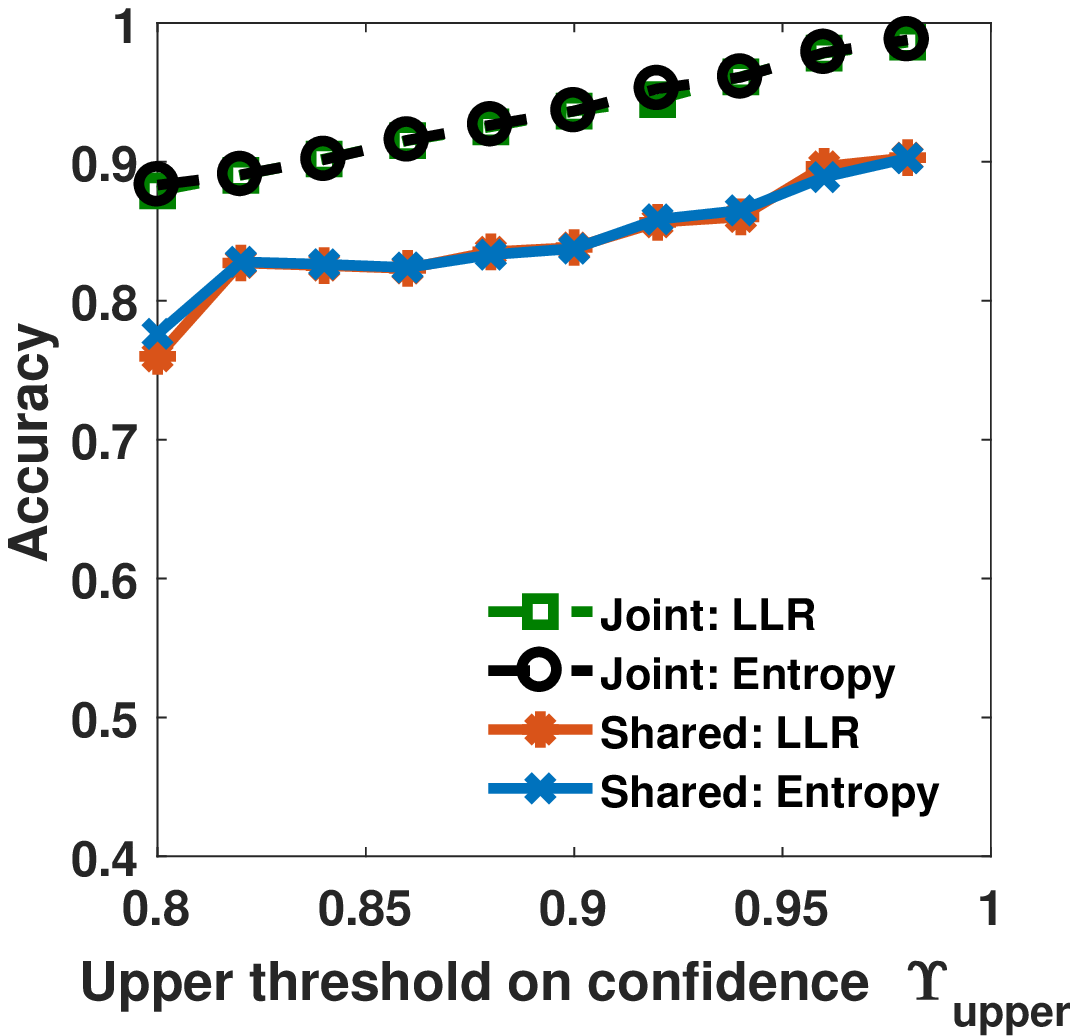}
\includegraphics[height= 5.4cm]{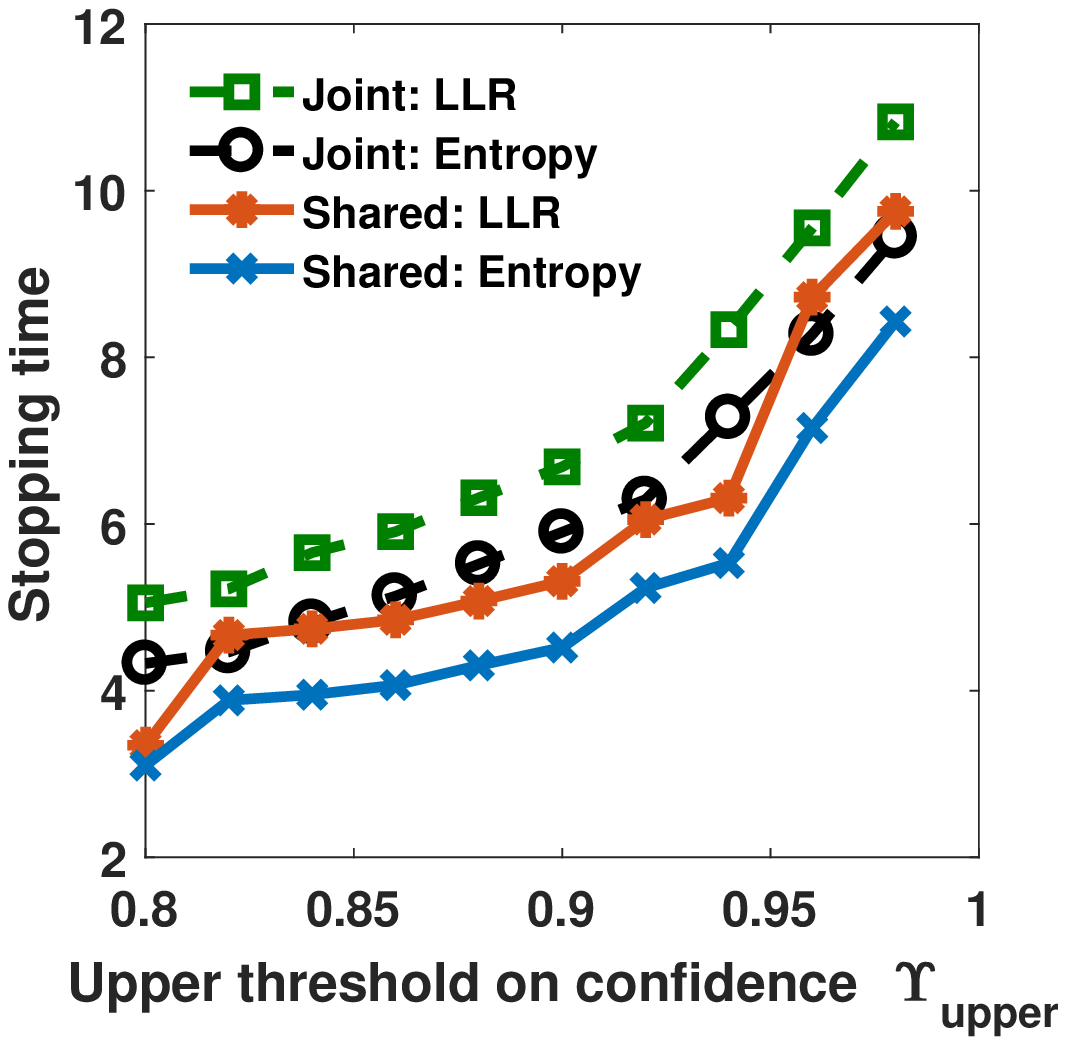}
\end{center}
\caption{Performances of the shared and joint detection algorithms as a function of the upper threshold on confidence $\upi$ when sensing cost per observation $\lambda=5$ and correlation coefficient $\rho=0.8$.}
\label{fig:Joint}
\end{figure*}

\subsection{Decentralized Algorithm} \label{sec:decentral_sim}
The architecture and parameters for our decentralized algorithm are as follows. We implement the actor and critic neural networks with four layers and the ReLU activation function between consecutive layers. The output layer of the actor layer uses the sigmoid function to ensure that the entries of the output vector of probabilities belong to set $[0,1]$. The parameters of the neural networks are updated using Adam Optimizer. We set the learning rates of the actor for the entropy and LLR-based reward functions as $2\times 10^{-5}$ and $3\times 10^{-5}$, respectively. Also, the critic learning rate is $1\times 10^{-4}$ for both reward functions. Additionally, we set the discount factor as $\gamma = 0.9$ and the regularizers as $\eta=1$ for the LLR-based reward function in \eqref{eq:reward_llr} and $\eta=0.1$ for the entropy-based reward function in \eqref{eq:reward_entropy}.

The numerical results for the decentralized algorithm under two settings are provided in \Cref{fig:Decentralized,fig:Joint}. In the first setting, all the sensors share their observations with all the other sensors, i.e., $\mathcal{N}_i=\{1,2,\ldots,N\}, \forall i$, which we refer to as the \emph{shared detection algorithm}. In the second setting, none of the sensors share their observations with any other sensor, i.e., $\mathcal{N}_i=\{i\}, \forall i$, which we refer to as the \emph{local detection algorithm}. We also consider a scheme with the joint pmf-based stopping rule (referred to as \emph{joint detection algorithm}). The joint detection algorithm is identical to the shared detection algorithm, but its stopping rule is based on the joint posterior probabilities. This method computes the joint posterior probabilities $\pi(k)\in[0,1]^{2^N}$ of all the entries of $\vecs\in[0,1]^N$ (similar to the algorithm in \cite{joseph2020anomaly}). The stopping rule based on joint pmf is $\underset{r=1,2,\ldots,2^N}{\max}\pi_r(k)>\upi$ and if the stopping rule is satisfied, the algorithm declares the state corresponding to $\underset{r=1,2,\ldots,2^N}{\arg\max}\pi_r(k)$ as $\hat{\vecs}$.

The performances of the shared and local detection algorithms are depicted in \Cref{fig:Decentralized}, and \Cref{fig:Joint} that provide comparisons of the shared and joint detection algorithms. From the first column of \Cref{fig:Decentralized,fig:Joint}, we first observe that the accuracy of all the algorithms increases with $\upi$, as expected. However, the accuracy is insensitive to $\lambda$ and $\rho$ because the accuracy primarily depends on the stopping rule that is independent of $\lambda$ and $\rho$. Since the  computations of the confidence used by the stopping rule are identical for the two versions (entropy and LLR-based schemes) of each algorithm (shared, local, and joint detection algorithms), they have the same accuracy levels. We see that the accuracy levels of the local and joint detection algorithms are slightly higher than our shared detection algorithm, but higher accuracies are obtained at the cost of a higher stopping time.

The middle subfigure of \Cref{fig:Decentralized}a shows that, unlike the accuracy, the number of observations per unit time is insensitive to the $\upi$. This trend is because the number of observations per unit time only depends on the selection policy learned by the algorithm, which in turn, depends only on the correlation coefficient $\rho$ and sensing cost per observation $\lambda$. We see that the number of observations per unit time decreases with $\lambda$. This behavior naturally follows from \eqref{eq:reward_decentral} because the second term corresponding to the number of observations in \eqref{eq:reward_decentral} gets more (negative) weight in the reward function compared to the first term (entropy or LLR term). We also note that the number of observations per unit time does not show any noteworthy change as  $\rho$ varies. Further, due to the common centralized training, the selection policies of all the algorithms with a common reward function are the same. Hence,  all the algorithms with a common reward function have the same number of observations per unit time.

The most sensitive performance metric is the stopping time which depends on both policy and stopping rule. The variation of stopping time as a function of different parameters is shown in the last column of \Cref{fig:Decentralized,fig:Joint}. The stopping times of all the algorithms increase with $\upi$ and $\lambda$. For a given policy, larger $\upi$ implies a higher accuracy level which leads to a longer stopping time. Similarly, a large value of $\lambda$ results in a small number of observations per unit time, and consequently, stopping time increases with $\lambda$ for any given $\upi$ and $\rho$. Further, the dependence of the stopping time on $\rho$ depends on the specific scheme. The local detection algorithm  ignores the correlation between the processes, and as a consequence, its stopping time does not vary significantly with $\rho$. However, the shared and joint detection algorithms update the marginal probabilities by accounting for the statistical dependence among the processes. As $\rho$ increases, each observation from a process gives more information on the corresponding correlated process and leads to a shorter stopping time.

Comparing the entropy-based and LLR-based algorithms, we see that both schemes offer almost the same level of accuracies and observations per unit time. However, the stopping time corresponding to the entropy-based scheme shows an improvement over the LLR-based schemes.

\section{Conclusion}\label{sec:conclusion}
We presented low-complexity centralized and decentralized algorithms to detect anomalous processes among a set of binary processes by dynamically selecting processes to be observed. The sequential process selection problem was formulated utilizing an MDP whose reward is defined using the entropy or LLR of the marginal probabilities of the states of the binary processes. The optimal process selection policy was obtained via the deep actor-critic RL algorithm that maximizes the long-term average reward of the MDP. The centralized algorithms were designed to choose one process per unit time, whereas, for the decentralized algorithm, the number of observations per unit was controlled by the sensing cost incorporated into the MDP reward. With the numerical results, we have analyzed the performance of our centralized and decentralized algorithms.
The results showed that our algorithms offered good detection accuracy and stopping time while operating with low complexity. The algorithms also exploited the underlying statistical dependence among the binary processes, which led to a shorter stopping time when the processes were highly correlated.
However, for scalable computing, our algorithms rely on approximate marginal probabilities. The approximation error depends on the underlying statistics and can lead to performance deterioration.  A theoretical analysis that quantifies the approximation error is an interesting direction for future work. 

\bibliographystyle{IEEEtran}
\bibliography{AnomalyDetection}

\end{document}